\documentclass{article}
\usepackage[ruled,vlined,linesnumbered]{algorithm2e}
\usepackage{hyperref}
\usepackage{url}
\usepackage{amssymb}
\usepackage{multirow}
\usepackage{float}
\usepackage[most]{tcolorbox}
\usepackage{adjustbox}
\usepackage{amsmath}
\usepackage{makecell}
\usepackage[table]{xcolor} 
\usepackage{booktabs}      
\usepackage{colortbl}
\definecolor{LightOrange}{rgb}{1,0.85,0.8}
\definecolor{blond}{rgb}{0.98, 0.94, 0.75}
\definecolor{blizzardblue}{rgb}{0.67, 0.9, 0.93}
\definecolor{LightGreen}{rgb}{0.93,0.98,0.96}
\definecolor{babypink}{rgb}{0.96, 0.76, 0.76}
\definecolor{classicrose}{rgb}{0.98, 0.8, 0.91}
\definecolor{textboxblue}{RGB}{22, 98, 132}
\definecolor{textboxgrey}{RGB}{252, 252, 252}
\definecolor{commentgreen}{RGB}{25, 107, 36}
\definecolor{lightblue}{rgb}{0.18,0.45,0.71} 
\usepackage{xcolor}
\SetKwProg{Fn}{Function}{}{end}

\SetKwFunction{FMain}{SIDiffAgent}
\SetKwFunction{FCreativity}{CreativityAnalysis}
\SetKwFunction{FIntent}{IntentionAnalysis}
\SetKwFunction{FRefine}{PromptRefinement}
\SetKwFunction{FNeg}{AdaptiveNegativePrompts}
\SetKwFunction{FGenerate}{GenerateOrEdit}
\SetKwFunction{FEval}{EvaluateImage}
\SetKwFunction{FRetrieve}{RetrieveGuidance}
\SetKwFunction{FInject}{InjectGuidance}
\SetKwFunction{FRecord}{RecordTrajectory}
\SetKwFunction{FUpdateKB}{UpdateKnowledgeBase}

\usepackage{booktabs}

\usepackage[preprint]{icml2026}

\usepackage{amsmath}
\usepackage{amssymb}
\usepackage{mathtools}
\usepackage{amsthm}

\usepackage[capitalize,noabbrev]{cleveref}

\usepackage[textsize=tiny]{todonotes}

\icmltitlerunning{SIDiffAgent: Self-Improving Diffusion Agent}
\usepackage{xspace}
\usepackage{listings}
\usepackage{tcolorbox}
\tcbuselibrary{listings, skins, breakable}

\lstdefinestyle{prompt}{
  language=bash,
  basicstyle=\small\ttfamily,
  backgroundcolor=\color{gray!10},
  frame=none,
  numbers=none,
  columns=fullflexible,
  breakatwhitespace=false, 
  breaklines=true,
  showstringspaces=false,
  xleftmargin=0pt,
  xrightmargin=0pt,
  aboveskip=0pt,
  belowskip=0pt,
  literate={~} {$\sim$}{1},
  escapeinside={(*@}{@*)}
}

\newtcolorbox{promptbox}[1][]{
  enhanced,
  breakable,
  colback=gray!10,     
  colframe=black,      
  colbacktitle=black,  
  coltitle=white,      
  title={#1},          
  boxrule=0pt,         
  top=0pt, bottom=0pt, left=0pt, right=0pt,
  listing only,
  listing options={style=prompt},
  sharp corners,
  fonttitle=\bfseries,
}

\newcommand\modelour{\emph{SIDiffAgent}\xspace}

\begin{document}

\twocolumn[{%
\renewcommand\twocolumn[1][]{#1}%
\icmltitle{SIDiffAgent: Self-Improving Diffusion Agent}

\icmlsetsymbol{equal}{*}

\begin{icmlauthorlist}
    \icmlauthor{Shivank Garg}{equal,iitr}
    \icmlauthor{Ayush Singh}{equal,iitr}
    \icmlauthor{Gaurav Kumar Nayak}{iitr}
\end{icmlauthorlist}

\icmlaffiliation{iitr}{Indian Institute of Technology, Roorkee}

\icmlcorrespondingauthor{Shivank Garg}{shivank\_g@mfs.iitr.ac.in}

\icmlkeywords{Machine Learning, ICML}
  \vskip 0.3in

\centering
\includegraphics[width=0.85 \textwidth]{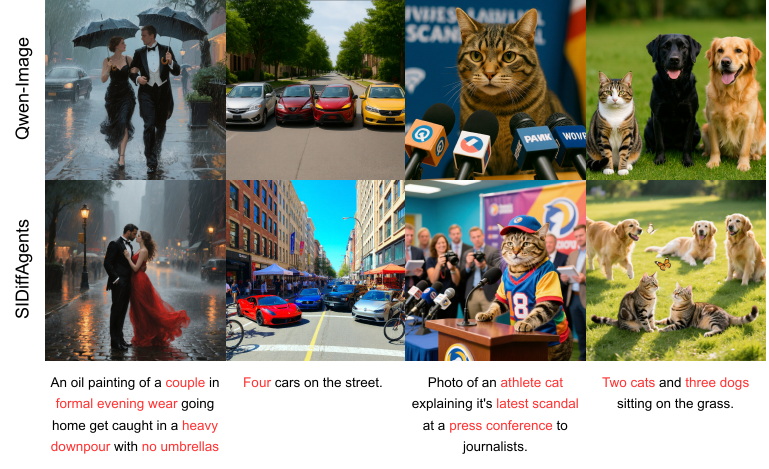} 
\vspace{-0.5em}
\captionof{figure}{Qualitative comparison between our model, \modelour, and the state-of-the-art, Qwen-Image model. \modelour demonstrates superior text-to-image alignment and realism, accurately rendering prompts that require nuanced understanding. Notably, our model successfully handles: negative constraints (`no umbrellas'), complex scene composition, anthropomorphic roles (`athlete cat'), and precise object counting.}
\label{fig:sample}
\vspace{1em}
}
]



\printAffiliationsAndNotice{\icmlEqualContribution}

\begin{abstract}
Text-to-image diffusion models have revolutionized generative AI, enabling high-quality and photorealistic image synthesis. However, their practical deployment remains hindered by several limitations: sensitivity to prompt phrasing, ambiguity in semantic interpretation (e.g., ``mouse" as animal vs. a computer peripheral), artifacts such as distorted anatomy, and the need for carefully engineered input prompts. Existing methods often require additional training and offer limited controllability, restricting their adaptability in real-world applications. We introduce Self-Improving Diffusion Agent (\modelour), a training-free agentic framework that leverages the Qwen family of models (Qwen-VL, Qwen-Image, Qwen-Edit, Qwen-Embedding) to address these challenges. \modelour autonomously manages prompt engineering, detects and corrects poor generations, and performs fine-grained artifact removal, yielding more reliable and consistent outputs. It further incorporates iterative self-improvement by storing a memory of previous experiences in a database. This database of past experiences is then used to inject prompt-based guidance at each stage of the agentic pipeline. \modelour achieved an average VQA score of 0.884 on GenAIBench, significantly outperforming open-source, proprietary models and agentic methods.
\end{abstract}

\begin{figure*}[!t]
    \centering
    \includegraphics[width=0.80\textwidth]{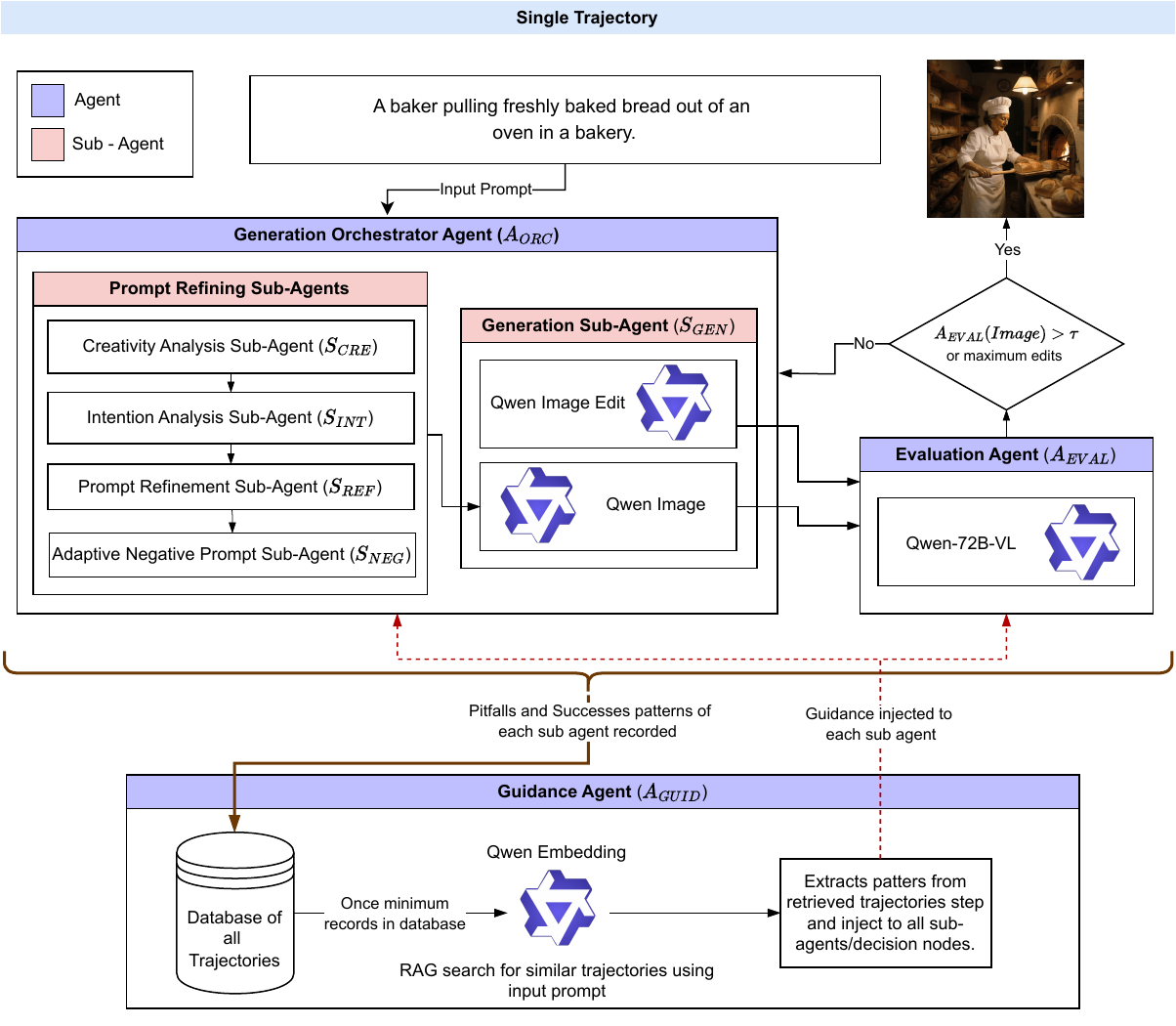}
    \caption{Workflow of the Self-Improving Diffusion Agent (\modelour). An input prompt is processed by the Generation Orchestrator Agent ($A_{\text{ORC}}$), which employs sub-agents ($S_{\text{CRE}}$, $S_{\text{INT}}$, $S_{\text{REF}}$, $S_{\text{NEG}}$) to assess creativity, clarify intent, refine the prompt, and add adaptive negative constraints before generation ($S_{\text{GEN}}$). The Evaluation Agent ($A_{\text{EVAL}}$) scores the generated image on aesthetic quality and text–image alignment, triggering targeted refinements if the evaluation score is less than the pre-defined threshold $\tau$ . Each trajectory is stored in the knowledge base, where the Guidance Agent ($A_{\text{GUID}}$) stores pitfalls and successes into corrective and workflow guidance, which are injected back into decision nodes to improve future generations.}
    \label{fig:approach_diagram}
\end{figure*}

\section{Introduction}
\label{sec:intro}
Image generation models ~\citep{rombach2021high,ramesh2022hierarchical,saharia2022photorealistic,singer2022make,ho2022imagen,blattmann2023align, esser2023structure} have transformed digital image synthesis, enabling the creation of high-quality visuals with remarkable detail and realism from simple text descriptions. Despite these advances, the performance of state-of-the-art text-to-image diffusion models remains highly sensitive to the precise phrasing of input prompts, making them unable to capture the true intent of the user and creating a significant ``intent gap'' between the user's intended meaning and the textual instructions provided to the model~\citep{chen2025t2i}. Here, intent refers to the underlying concept or goal that the user wishes the model to capture. This gap arises because natural language prompts are often ambiguous or casually phrased, leading the model to generate outputs that deviate from the user's goal. For instance, the prompt ``a mouse" may yield different outputs, either an animal or a computer peripheral, depending on how the model interprets the text.

Another key challenge lies in the distributional mismatch between the unstructured, free-form language typical of end-user prompts and the highly curated, descriptive captions created by human experts and subjected to intensive filtering during the training of diffusion models~\citep{li2024hunyuan,wu2025qwen,flux24}. This discrepancy often results in inadequate textual guidance for generation. For instance, prompts are frequently underspecified (e.g., ``a car on a road" without details on the car's model, color, or environment), which forces the model to make unguided assumptions. Such underspecification not only compromises output quality but also increases computational costs, as users must repeatedly regenerate outputs to achieve satisfactory results. These issues constitute a critical bottleneck for the scalability and reliability of diffusion models in professional domains where robust and predictable alignment with user intent is essential.

To mitigate these challenges, prior work such as T2I-Copilot~\citep{chen2025t2i} and MCCD~\citep{li2025mccd} have proposed agentic frameworks that address ambiguities through image editing and agentic coordination. However, these approaches often lack fine-grained controllability and robust mechanisms for post-generation artifact correction. A key challenge in such Multi-Agent Systems is coordination. Agents often have only partial observability and limited knowledge of others capabilities or the effects of their own actions, leading to suboptimal outcomes \citep{cemri2025multi}. Recent works in the field of LLM Agents address this via Theory of Mind (ToM), the ability to attribute goals and beliefs to others \citep{cross2024hypotheticalmindsscaffoldingtheory, yang2025largelanguagemodelstheory}. With ToM, agents can understand peers intentions and anticipate future actions, enabling better coordination. To the best of our knowledge, no work has applied this to diffusion-based agents. 

Additionally, ReNeg~\citep{li2025reneg} highlights the role of negative prompts and embeddings~\citep{ho2022classifier} in constraining diffusion processes and guiding models away from undesired artifacts. In this work, we introduce a training-free, multi-agent framework \modelour that creates a robust and self-improving Theory of Mind inspired image generation pipeline. \modelour leverages the capabilities of the Qwen model family (Qwen-VL~\citep{Qwen2.5_VL}, Qwen-Image~\citep{wu2025qwen}, Qwen-Edit~\citep{wu2025qwen}, Qwen-Embedding~\citep{zhang2025qwen3embeddingadvancingtext}) as the backbone of our agentic flow. Our framework uses multiple agents to refine prompts, reuse past experience, and fix artifacts with local edits. Further building upon the observations from ReNeg~\citep{li2025reneg}, \modelour also incorporates an additional adaptive negative prompt generation agent. Beyond controllability and artifact mitigation, we emphasize the importance of self-improvement: unlike prior agentic frameworks, \modelour is designed to leverage past successes and failures to iteratively refine its generation strategies. While our contribution focuses on system-level architecture rather than a new training objective, we highlight three technically novel components. First, \modelour introduces the first experience-driven memory mechanism for diffusion-based agents, enabling retrieval-conditioned guidance throughout the generation workflow. Second, our Theory-of-Mind–inspired inter-agent reasoning models predictive expectations over other agents' behaviors using accumulated success/pitfall patterns, providing dynamic workflow guidance beyond prompt optimization. Third, SIDiffAgent offers a training-free alternative to RL-based or reward-guided finetuning approaches (e.g., DPO-Diffusion, ReNeg), enabling practical quality improvements even for closed-weight diffusion systems without any retraining.




We empirically validate the effectiveness of \modelour across multiple benchmarks, including GenAIBench and DrawBench, where it establishes new state-of-the-art performance. Our approach not only surpasses prior agentic systems such as T2I-Copilot, achieving a VQA Score~\citep{VQAScore_ECCV24} improvement of 8.73\%, but also outperforms leading proprietary models (e.g., Recraft V3, Imagen 3, Flux 1.1-Pro), delivering a 5.36\% improvement over Imagen 3. Furthermore, it significantly exceeds the performance of powerful open-source systems (e.g., Flux.1-Dev, HunyuanDiT v1.2, and SD 3.5), achieving a 15.70\% improvement over SD 3.5.

Our key contributions are summarized as follows:
    \begin{itemize}    
    \item We introduce a multi-agent system that refines the inputs to a diffusion model at test time, ensuring consistent understanding of prompts even when they are expressed in different ways. This results in generations that more faithfully capture the user's intended concepts.

    \item We adapt a structured image editing approach that enables targeted, incremental corrections to generated images without requiring complete regeneration, narrowing the intent gap and improving alignment between user goals and model generations.

    \item We build an iterative self-improvement mechanism in which an agentic memory tracks successes, artifacts and pitfalls of various agents, and using them to inject corrective guidance based on Theory of Mind. This creates a training-free, self-improving agentic framework that progressively enhances generation quality over time.
\end{itemize}


\section{Related Works}
\label{sec:related}

\subsection{AI Agents} 
Agents are autonomous entities capable of perceiving their environment, reasoning over goals, and executing actions, often collaboratively through structured memory, planning, and tool use~\citep{luo2025large}. In commercial applications such as browser automation or coding~\citep{hong2024metagpt,qian2023chatdev}, these agents typically rely on a large language model (LLM) backbone, which performs reasoning over the current state to determine the optimal action. An extension of this paradigm is the multi-agent framework, in which multiple agents collaborate not only to reason over their own states but also to reason about the states of other agents~\citep{li2023camel}. Building on these foundations, researchers have started exploring agentic workflows for image generation. In such systems, multi-agent pipelines decompose and refine prompts, evaluate outputs, and correct artifacts. Examples include T2I-Copilot~\citep{chen2025t2i}, which integrates prompt interpretation, model selection, and quality evaluation, and PromptSculptor~\citep{xiang2025promptsculptor}, which iteratively transforms vague user prompts into precise ones. These systems demonstrate how agentic principles can reduce the intent gap in text-to-image generation, improving fidelity, controllability, and efficiency without requiring additional model training.
\subsection{Theory of Mind in Agentic Systems}
Building on multi-agent systems, Theory of Mind (ToM)~\citep{rabinowitz2018machine,rocha2023applying} has become an increasingly important concept for designing smarter and more adaptive agents. It allows agents to understand, infer, and reason about what other agents might believe, desire, or intend in a given context. Recent research in multi-agent and conversational settings has shown that endowing agents with this capability significantly improves both overall system performance and individual agent effectiveness~\citep{yang2025large,gu2025agentgroupchat,jim2001communication}. To the best of our knowledge, this idea has not yet been systematically applied to diffusion-based generative models. In \modelour, ToM is implemented through specialized sub-agents that continuously track each other's mental states, typical successes, potential pitfalls, and characteristic behavior patterns. This information is integrated with a self-improvement mechanism that learns from past experiences with similar inputs to dynamically update these states over time. As a result, each sub-agent can proactively anticipate potential pitfalls and successes of other agents, adjusting its own reasoning accordingly. This awareness allows agents to make decisions that improve not only their individual output quality but also the collective performance of the other sub-agents, ultimately leading to a more robust, adaptive, and coordinated agentic system.

\subsection{Prompt Optimization for Diffusion Models}

Prompt Optimization (PO) has emerged as a crucial technique for aligning large language models (LLMs) with human intent and improving downstream task performance. Originating from in-context learning, PO has evolved into automatic prompt engineering, where LLMs iteratively refine and adapt prompts to achieve more precise and contextually appropriate outputs. While PO has been extensively studied in the context of LLMs, its application to diffusion models remains comparatively underexplored, despite the growing importance of text-to-image generation. For instance, Promptist~\citep{hao2023optimizing} fine-tunes an LLM via reinforcement learning from human feedback to augment user prompts with stylistic and artistic modifiers, while OPT2I~\citep{manas2024improving} leverages an LLM to refine prompts directly for higher-quality image synthesis. Other works, such as DPO-Diffusion~\citep{wang2024discrete} and ReNeg~\citep{li2025reneg}, focus on optimizing negative prompts, either through LLM reasoning or learned embeddings that guide the generative process away from undesired features. TextGrad~\citep{yuksekgonul2024textgrad} introduces gradient-based text feedback optimization, and manual strategies for prompt engineering have also been systematically investigated. Our agentic system builds on these ideas by jointly optimizing positive and image-specific negative prompts at test time via prompt engineering, dynamically balancing guidance to produce more reliable, semantically aligned, and visually coherent images without requiring additional model training.

\subsection{Self-Improving AI} Recent work has also explored self-improving models and agents, where systems enhance their own performance via interaction or feedback. For LLMs, approaches such as LMSI~\citep{huang2024self} leverage reward signals for autonomous improvement, while IPO~\citep{garg-etal-2025-ipo} and Self-Rewarding~\citep{yuan2024self} LLMs optimize through preference pair construction. Methods like SPPO~\citep{wuself} apply self-play to iteratively refine capabilities. Beyond LLMs, agentic systems such as MaestroMotif~\citep{klissarovmaestromotif} use feedback-driven skill acquisition, and AgentSquare~\citep{shang2024agentsquare} abstracts agent design into modular components for automated exploration. Other works, such as AlphaEvolve~\citep{novikov2025alphaevolvecodingagentscientific} and ADAS~\citep{hu2025automateddesignagenticsystems}, employ meta-agents to improve task-specific agents, while SICA~\citep{robeyns2025selfimprovingcodingagent} eliminates the meta/target distinction by enabling agents to directly edit their own codebases. While the field of self-improving LLMs and coding agents has been actively studied, diffusion-based agents have not been explored in this context. To the best of our knowledge, we present the first training-free, self-improving diffusion framework, where agentic memory and prompt engineering enable self-improvement without retraining.

\section{Proposed Method: Self-Improving Diffusion Agent (\modelour)}

Our framework is built around an agentic workflow that coordinates prompt engineering, image generation, evaluation, and self-improvement. This workflow is orchestrated by three primary agents, the Generation Orchestrator Agent ($A_{\text{ORC}}$), which preprocesses user prompts through four sub-agents, the Creativity Analysis Sub-Agent ($S_{\text{CRE}}$), Intention Analysis Sub-Agent ($S_{\text{INT}}$), Prompt Refinement Sub-Agent ($S_{\text{REF}}$), and Adaptive Negative Prompt Sub-Agent ($S_{\text{NEG}}$) and Generation Sub-Agent ($S_{\text{GEN}}$) for the final generation. The Evaluation Agent ($A_{\text{EVAL}}$) assesses each output, triggering editing when necessary. Finally, the Guidance Agent ($A_{\text{GUID}}$) facilitates learning from past experiences. The overall workflow of \modelour is depicted in Figure~\ref{fig:approach_diagram} and the algorithm is given in Appendix~\ref{algo}. Each agent is discussed in detail below.

\subsection{Generation Orchestrator Agent ($A_{\text{ORC}}$)}
The Generation Orchestrator Agent forms the foundation of our approach. It performs a structured preprocessing sequence that balances user intent, creative enrichment, and model compatibility. To achieve this, it operates through five specialized sub-agents, each addressing a distinct aspect of prompt engineering and generation. The following is a detailed explanation of all the agents. The prompts for each of the agents are in the Appendix~\ref{orch}.

\textbf{Step 1. Creativity Analysis Sub-Agent ($S_{\text{CRE}}$) :} The first step is to estimate the freedom that the agents can have which guides the intensity of refinements in the later stages. It classifies the input into three levels: high for short or vague prompts, requiring substantial enrichment; medium for moderately detailed prompts, needing balance between structure and interpretation; and low for highly detailed prompts, where strict adherence is key. For instance, the prompt `a cat' is classified as high creativity, (e.g., specifying breed, pose, or environment), whereas the prompt `a Persian cat sitting on a red sofa in a living room' would be classified as low creativity, since it already provides detailed constraints. Figure~\ref{fig:creative} showcases the prompts and image generation pairs for different creativity levels.

\textbf{Step 2. Intention Analysis Sub-Agent ($S_{\text{INT}}$) :} After establishing the creativity level, this sub-agent deconstructs the users input to produce a comprehensive semantic layout of the prompt. The process extracts core content from the prompt, such as \textit{main subjects}, their \textit{attributes}, their \textit{spatial relationships}, \textit{background}, and additional parameters like \textit{composition}, \textit{lighting}, and overall \textit{visual style}.

Beyond structural parsing, another critical function of this agent is to identify and resolve ambiguities or missing information within the prompt conditioned on the creativity level determined by ($S_{\text{CRE}}$). For prompts designated as low creativity, the agent generates minor assumptions. Conversely, for high creativity prompts, it autonomously assumes information while adhering to the input prompt and ensuring coherent and contextually appropriate details. The output of this stage is a structured specification that formalizes the user's intent and generates clarifications. For example, given the prompt `a car' at low creativity, $S_{\text{INT}}$ preserves the minimal description without adding details. At high creativity, however, it expands the prompt by generating clarifying attributes such as the car model, color, etc. These details are autonomously inferred by $S_{\text{INT}}$ to maintain coherence while enriching the original input.

\textbf{Step 3. Prompt Refinement Sub-Agent ($S_{\text{REF}}$) :} Building upon the semantic layout generated from $S_{\text{INT}}$, $S_{\text{REF}}$ produces a clearer and refined version of the prompt. $S_{\text{REF}}$ ensures fidelity to the user's intent by preserving all original subjects and scene elements, while simultaneously restructuring the prompt into a more contextually rich and coherent form.

\textbf{Step 4. Adaptive Negative Prompt Sub-Agent ($S_{\text{NEG}}$) :} In classifier-free guidance in diffusion models, the model normally contrasts the conditional prompt with an unconditional prompt~\citep{ho2022classifier}. Negative prompts are injected by replacing that unconditional branch with the negative text embedding, so the guidance steers outputs away from undesired features. This provides fine-grained control, enabling the explicit exclusion of unwanted objects, styles, or quality defects. While the refined prompt captures what should appear in the image, an equally important step is to constrain what must not appear. 

$S_{\text{NEG}}$ first appends universal safeguards to mitigate common artifacts (e.g., `low quality, blurry, distorted, watermark'). Then it analyzes the refined prompt to derive semantically relevant, scene-specific negations. For instance, a prompt specifying a ``clear blue sky" adds ``clouds, dark clouds" to the negative prompt, while ``a person standing alone" generates an exclusion for ``extra people or crowd". These adaptive constraints are concise, context-aware, thereby improving output quality and adherence without undermining user intent.

\textbf{Step 5. Generation Sub-Agent ($S_{\text{GEN}}$) :} The final component in $A_{\text{ORC}}$ is the generation sub-agent, which is responsible for the task of generating the image. This sub-agent has 2 models (Qwen-Image and Qwen-Image-Edit) to handle both initial creation and subsequent refinement. Initially, it employs Qwen-Image to generate an image from the refined positive and adaptive negative prompts. Following this step, the evaluation is initiated (Discussed in Section~\ref{subsec:eval}). If $A_{\text{EVAL}}$ identifies defects or misalignments, $S_{\text{GEN}}$ invokes Qwen-Image-Edit, a specialized editing model, to perform targeted corrections. This strategy of combining generation with precise editing allows the system to iteratively improve the output, ensuring higher fidelity to complex user specifications. Appendix~\ref{fig:creative} showcases the targeted editing by Qwen-Image-Edit.

\begin{table*}[!t]
    \begin{center}
    \small
    \scalebox{0.75}{%
    \setlength{\tabcolsep}{1.5pt}
    \begin{tabular}{lccccccccccccc|c}
    \toprule
    \multirow{3}{*}{Method} 
    & \multicolumn{13}{c}{GenAI-Bench} 
    & \multirow{3}{*}{DrawBench} \\

    \cmidrule(lr){2-14}

    & \multicolumn{6}{c}{Basic} 
    & \multicolumn{6}{c}{Advanced} 
    & \multirow{2}{*}{Overall} \\

    \cmidrule(lr){2-7}\cmidrule(lr){8-13}

    & \multirow{2}{*}{Attribute} 
    & \multirow{2}{*}{Scene} 
    & \multicolumn{3}{c}{Relation} 
    & \multirow{2}{*}{Overall} 
    & \multirow{2}{*}{Count} 
    & \multirow{2}{*}{Differ} 
    & \multirow{2}{*}{Compare} 
    & \multicolumn{2}{c}{Logical} 
    & \multirow{2}{*}{Overall} 
    & \\

\cmidrule(lr){4-6}\cmidrule(lr){11-12}

        & & & Spatial & Action & Part & & & & & Negate & Universal & & & \\
        \midrule
        \rowcolor[HTML]{EFEFEF} 
        \textit{Proprietary} & & & & & & & & & & & & & & \\
        Imagen 3 v002~\citep{imagen3_24} & 0.909 & 0.923 & 0.909 & 0.903 & 0.918 & 0.912 & 0.841 & 0.841 & 0.795 & 0.673 & 0.788 & 0.776 & 0.839 & \underline{0.866}\\
        \scriptsize{(Task completion rate)} & \scriptsize{(92.4\%)} & \scriptsize{(93.1\%)} & \scriptsize{(93.0\%)} & \scriptsize{(89.3\%)} & \scriptsize{(88.7\%)} & \scriptsize{(92.3\%)} & \scriptsize{(91.5\%)} & \scriptsize{(88.9\%)} & \scriptsize{(90.7\%)} & \scriptsize{(88.8\%)} & \scriptsize{(89.1\%)} & \scriptsize{(90.7\%)} & \scriptsize{(91.4\%)} & \scriptsize{(97.0\%)}\\
        Recraft v3~\citep{recraftv3_24} & 0.914 & 0.913 & 0.913 & 0.901 & 0.913 & 0.913 & 0.806 & 0.797 & 0.772 & 0.589 & 0.761 & 0.725 & 0.811  & 0.836\\
        FLUX1.1-pro~\citep{flux24} & 0.890 & 0.899 & 0.884 & 0.871 & 0.894 & 0.884 & 0.766 & 0.788 & 0.751 & 0.490 & 0.710 & 0.666 & 0.766 & 0.786\\
        Midjourney v6~\citep{midjourneyv6.1_24} & 0.880 & 0.870 & 0.870 & 0.870 & 0.910 & 0.870 & 0.780 & 0.780 & 0.790 & 0.500 & 0.760 & 0.690 & 0.772 & -\\
        DALL-E 3~\citep{dalle3_24} & 0.910 & 0.900 & \underline{0.920} & 0.890 & 0.910 & 0.900 & 0.820 & 0.780 & \underline{0.820} & 0.480 & \underline{0.800} & 0.700 & 0.791 & -\\
        \midrule
        \rowcolor[HTML]{EFEFEF} 
        \textit{Open-source} & & & & & & & & & & & & & & \\
        Kolors v1.0~\citep{kolors_arxiv24} & 0.821 & 0.841 & 0.832 & 0.818 & 0.803 & 0.819 & 0.737 & 0.726 & 0.705 & 0.438 & 0.695 & 0.621 & 0.711 & 0.646 \\
        Playground v2.5~\citep{playground_arxiv24} & 0.818 & 0.850 & 0.803 & 0.818 & 0.821 & 0.815 & 0.732 & 0.696 & 0.721 & 0.499 & 0.695 & 0.640 & 0.720 & 0.743 \\
        HunyuanDiT v1.2~\citep{hunyuandit_24} & 0.817 & 0.855 & 0.825 & 0.827 & 0.798 & 0.818 & 0.732 & 0.723 & 0.743 & 0.475 & 0.692 & 0.640 & 0.721 & 0.712\\
        Janus Pro-7B~\citep{janusPro7B_25} & 0.865 & 0.886 & 0.867 & 0.856 & 0.870 & 0.859 & 0.731 & 0.759 & 0.734 & 0.480 & 0.693 & 0.653 & 0.747 & 0.786\\
        Lumina-Image-2.0~\citep{luminaImage_25} & 0.879 & 0.896 & 0.876 & 0.872 & 0.885 & 0.874 & 0.760 & 0.767 & 0.729 & 0.451 & 0.723 & 0.649 & 0.752 & 0.790\\
        SD 3.5 large~\citep{SD35_24} & 0.891 & 0.895 & 0.889 & 0.880 & 0.895 & 0.890 & 0.760 & 0.763 & 0.743 & 0.471 & 0.707 & 0.659 & 0.764 & 0.781\\
        FLUX.1-dev~\citep{flux24} & 0.873 & 0.875 & 0.862 & 0.853 & 0.875 & 0.864 & 0.747 & 0.756 & 0.733 & 0.456 & 0.711 & 0.646 & 0.745 & 0.769\\
        GenArtist~\citep{Genartist_NeurIPS24} & 0.702 & 0.736 & 0.659 & 0.677 & 0.688 & 0.693 & 0.553 & 0.473 & 0.518 & 0.437 & 0.546 & 0.504 & 0.588 & 0.607 \\
        T2I-Copilot~\citep{chen2025t2i} & 0.893 & 0.909 & 0.893 & 0.885 & 0.899 & 0.892 & 0.813 & 0.807 & 0.759 & 0.659 & 0.766 & 0.747 & 0.813 & 0.829\\
        \midrule
        \rowcolor[HTML]{EFEFEF} 
        \textit{Qwen-Image-Based} & & & & & & & & & & & & & & \\
        Qwen-Image & 0.857 & 0.874 & 0.858 & 0.853 & 0.861 & 0.853 & 0.766 & 0.777 & 0.753 & 0.501 & 0.738 & 0.677 & 0.757 & 0.853\\
        Qwen-Agents & 0.906 & 0.912 & 0.910 & 0.897 & 0.921 & 0.907 & 0.790 & 0.814 & 0.777 & 0.497 & 0.742 & 0.690 & 0.789 & 0.833\\
        Qwen-Agent$_{neg}$ & 0.883 & 0.900 & 0.885 & 0.882 & 0.888 & 0.884 & 0.824 & 0.820 & 0.777 & 0.684 & 0.760 & 0.764 & 0.818 & 0.808 \\
        Qwen-Agents$_{vneg}$ & 0.913 & 0.912 & 0.913 & 0.905 & \underline{0.929} & 0.913 & 0.841 & 0.838 & 0.795 & 0.700 & 0.783 & 0.782 & 0.842 & 0.849 \\ 
        \modelour & \underline{0.919} & \underline{0.924} & \underline{0.918} & \underline{0.909} & 0.928 &\underline{0.918} & \underline{0.857} & \underline{0.855} & 0.805 & \underline{0.715} & 0.797 & \underline{0.796} & \underline{0.852} & 0.860\\   
        \emph{SIDiffAgent}$_{ep2}$ & \textbf{0.940} & \textbf{0.942} & \textbf{0.939} & \textbf{0.934} & \textbf{0.947} & \textbf{0.939 }& \textbf{0.890} & \textbf{0.879} & \textbf{0.833} & \textbf{0.771} & \textbf{0.837} & \textbf{0.836} & \textbf{0.884} & \textbf{0.901}\\

        \bottomrule
    \end{tabular}}
    \end{center}
    \caption{\textbf{Quantitative comparison of \modelour against 14 methods on DrawBench~\citep{Imagen1_NeurIPS22} and GenAI-Bench~\citep{GenAIBench_CVPRW24}}, evaluated using VQAScore~\citep{VQAScore_ECCV24}. The results for the baselines are taken from T2I-Copilot~\citep{chen2025t2i} and our generations were performed with identical random seeds ensuring fairness in evaluation. The best and second-best results are highlighted in \textbf{bold} and \underline{underlined}, respectively.}
    \label{tab:quan_t&i_alignment}
    \vspace{-0.3cm}
\end{table*}

\subsection{Evaluation Agent ($A_{\text{EVAL}}$)}
\label{subsec:eval}
$A_{\text{EVAL}}$ assesses the quality of each generated image and outputs a structured evaluation report. The assessment is based on two primary metrics: \textbf{Aesthetic Quality} and \textbf{Text–Image Alignment}, each scored on a 0–10 scale. The prompt used for $A_{\text{EVAL}}$ is in Appendix~\ref{eval_agent_p}.

Aesthetic Quality reflects the overall merit of the image, considering composition, color harmony, lighting, focus, sharpness, emotional impact, and uniqueness. Text–Image Alignment measures consistency with the prompt by verifying subject presence, spatial correctness, adherence to style, and background consistency. The overall performance score is computed as the average of the two scores. Beyond scoring, the agent also detects visual artifacts (e.g., noise, tiling, blending errors), identifies missing elements, and provides targeted suggestions for refinement. 

If the score falls below a pre-defined threshold, $S_{\text{GEN}}$ triggers an editing loop in which the image is edited via Qwen-Image-Edit using the suggested improvements by generating a refined prompt utilising $A_{\text{ORC}}$ based on suggestions generated by $A_{\text{EVAL}}$. This process repeats until the image reaches a satisfactory score or the maximum editing attempts are exhausted.

\subsection{Guidance Agent ($A_{\text{GUID}}$)}
The Guidance Agent enables the self-improving component of our workflow. A trajectory denotes a complete generation sequence, from the initial prompt to the final image, while decision nodes mark points where sub-agents perform key actions. $A_{\text{GUID}}$ records each trajectory in a structured format by condensing trajectories into pitfalls and successes for every decision node. These records accumulate in a persistent knowledge base, allowing the system to detect recurring patterns, such as specific sub-agents consistently failing on certain prompt types. When processing new inputs, $A_{\text{GUID}}$ retrieves relevant trajectories and provides both corrective guidance (derived from past pitfalls and successes) and workflow guidance (a structured description of the process) to each sub-agent. Our hypothesis is that this combination equips sub-agents with richer context and leads to more reliable decisions. The prompts used for $A_{\text{GUID}}$ are in Appendix~\ref{prompt_guidance_agent}

\textbf{1. Knowledge Base Construction:} The process begins by populating a `global memory', which serves as the system's persistent knowledge base. Each complete workflow execution, or `trajectory', is recorded for analysis. Instead of storing raw logs, a post-processing step occurs to analyze each trajectory. At each `decision node', it extracts structured insights by identifying specific `pitfalls' and `successes'. This compressed information is then stored in the knowledge base, which is structured with dedicated databases for different models (Qwen-Image and Qwen-Image-Edit) to facilitate model-specific pattern recognition.

\textbf{2. Guidance Formulation:} When a new user prompt is received, the Guidance Agent uses it as a query to retrieve the $K$-most semantically similar trajectories from the knowledge base via RAG. The agent then analyzes the structured data from these retrieved trajectories, aggregating the recorded pitfalls and successes associated with each decision node. By identifying high-frequency patterns such as a sub-agent repeatedly failing on a certain kind of prompt, it synthesizes a corrective guidance. This guidance contains actionable instructions, such as specific negative prompts to add or adjustments to be made, designed to mitigate known failure modes and replicate successful outcomes for the current task. Additionally, it also synthesizes a static workflow guidance, which provides each decision node about the overall workflow and its role in it. 

\textbf{3. Guidance Injection.} The synthesized guidance is then injected into the active workflow. At each decision node, the corresponding sub-agent receives both the static workflow guidance and the dynamic corrective guidance (formulated in the previous step) as part of its input context. The sub-agent then acts on this information. The static workflow guidance helps each decision node anticipate how its result not only affects the generation output but also the performance of other sub-agents. 
For instance, if the model is prompted to generate an image of a wall clock, ($A_{\text{GUID}}$) learns that ($S_{\text{GEN}}$) is unable to generate wall clocks that show a time other than 10:10~\citep{clock_1,clock_2}, and hence it would suggest ($A_{\text{ORC}}$) to explicitly guide the generation towards having the wall clock showing 10:10 and an overall low creativity.

The operation of $A_{\text{GUID}}$ can be understood in three stages: knowledge base construction, guidance formulation, and guidance injection. This cyclical process of retrieving guidance, applying it, and learning from the outcome forms a robust, training-free feedback loop. It is through this process the system continuously enhances its performance and adapts its strategies based on accumulated experience.
 
\section{Experiments}

\subsection{Experimentation Setup}

\begin{figure*}[!t]
    \centering
    \includegraphics[width=0.80\textwidth]{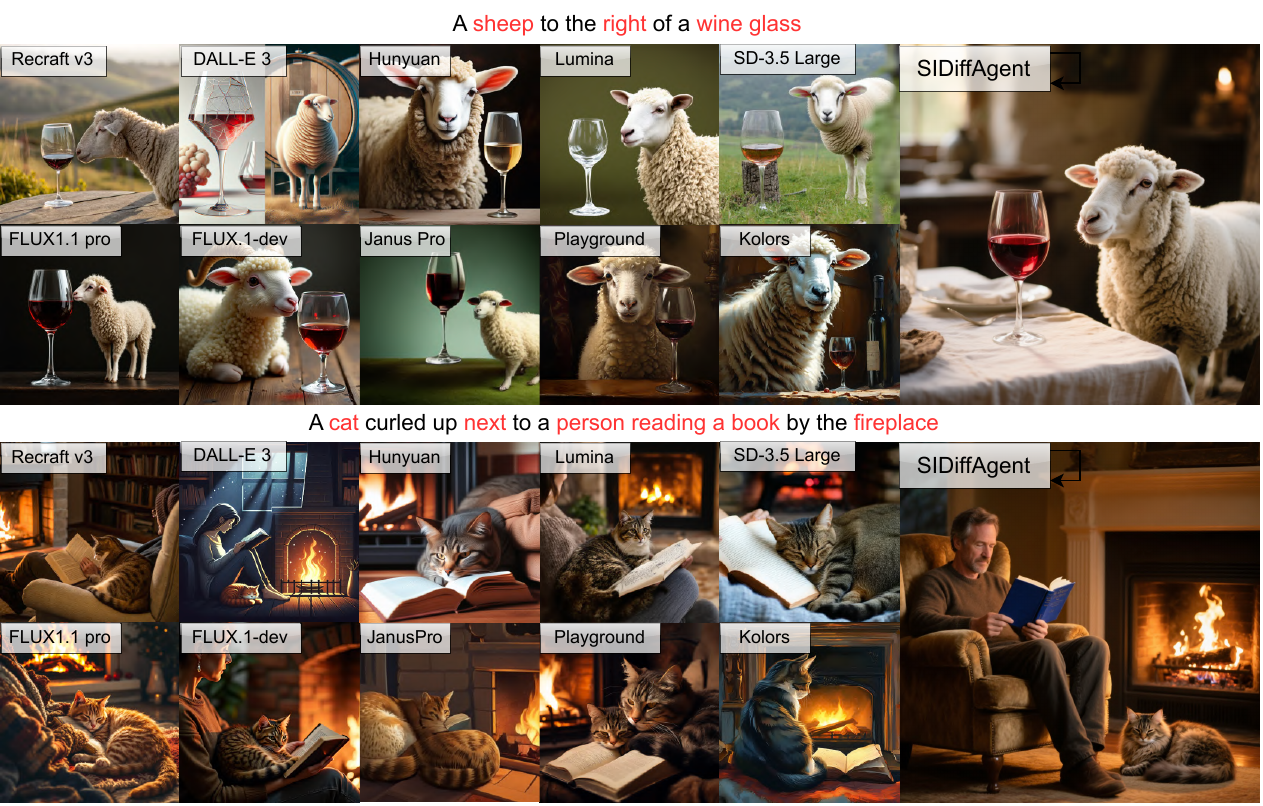} 
    \caption{\textbf{Qualitative comparison of \modelour with 10 models on prompts from DrawBench (Top) and GenAI-Bench (Bottom).} \modelour demonstrates superior handling of spatial relationships (correctly placing the sheep `to the right of' the glass) and compositional complexity (generating all elements of the cat, person, and fireplace scene) compared to other models.}
    \label{comparisons}
\end{figure*}

All experiments are conducted on a single node equipped with 8 $\times$ 80GB NVIDIA A100 GPUs. The Qwen-2.5-72B-VL\footnote{\url{https://huggingface.co/Qwen/Qwen2.5-VL-72B-Instruct}} model is hosted on 4 GPUs using vLLM~\citep{kwon2023efficient}, while the remaining 4 GPUs are allocated for image generation. The auxiliary models Qwen-Image~\citep{wu2025qwen}, Qwen-Edit~\citep{wu2025qwen}, and Qwen-Embedding~\citep{zhang2025qwen3embeddingadvancingtext} together require approximately 47GB of VRAM on a single GPU, with Qwen-Image and Qwen-Edit deployed under 4-bit quantization to reduce memory footprint and improve the speed. We built our multi-agent framework using LangGraph~\citep{langgraph}. During the image generation process, if a generated image received a score below 8.0 from $A_{\text{EVAL}}$, the system triggered editing, with a maximum of two attempts, following T2I-Copilot~\citep{chen2025t2i}.

The guidance system was implemented with a SQLite database storing condensed trajectories. Once 200 trajectories were accumulated the guidance generation activates. For each new prompt, we retrieved the top-$k=5$ most semantically similar trajectories using Qwen-Embedding-0.6B with FAISS~\citep{douze2024faiss}.

We evaluated the agentic system on two benchmarks, GenAI-Bench~\citep{GenAIBench_CVPRW24} and DrawBench~\citep{Imagen1_NeurIPS22}, using the generation seeds provided in the dataset. We employed VQAScore~\citep{VQAScore_ECCV24} as our primary evaluation metric for image quality and prompt alignment, which was identified as more human-aligned than CLIPScore~\citep{hessel2021clipscore}, PickScore~\citep{kirstain2023pick} and ImageReward~\citep{xu2023imagereward} by Imagen3~\citep{baldridge2024imagen}. We compare \modelour against proprietary models(Imagen~\citep{imagen3_24}, Recraft~\citep{recraftv3_24}, Flux-pro~\citep{flux24}, Midjourney~\citep{midjourneyv6.1_24}, and DALL-E~\citep{dalle3_24}), open-source models (Kolors~\citep{kolors_arxiv24}, Playground v2.5~\citep{playground_arxiv24}, Hunyuan DiT~\citep{hunyuandit_24}, Janus Pro~\citep{janusPro7B_25}, Lumina-Image 2.0~\citep{luminaImage_25}, SD 3.5~\citep{SD35_24}, Flux dev~\citep{flux24}, GenArtist~\citep{Genartist_NeurIPS24}, and Qwen Image~\citep{wu2025qwen}), and agentic approaches such as T2I-Copilot~\citep{chen2025t2i}.

\subsection{Ablation Studies} 
\label{sec:Ablation}
To understand the contribution of each component, we conduct five ablation studies:
\begin{itemize}
    \item \textbf{Qwen-Image:} Direct generations from Qwen-Image.
    \item \textbf{Qwen-Agents:} Full agentic workflow without adaptive negative prompts or guidance (comparable to T2I-Copilot~\citet{chen2025t2i}, but using Qwen-Image, Qwen-Image-Edit and Qwen-VL).
    \item \textbf{Qwen-Agent$_{neg}$:} Qwen-Agents with a fixed negative prompt (Appendix~\ref{const_negp}), showcasing the role of using negative prompts.
    \textbf{\item Qwen-Agent$_{vneg}$:} Qwen-Agents with variable negative prompt, showcasing the role of using variable negative prompts.
    \item \textbf{\modelour:} Our full framework with adaptive negative prompts and the guidance agent, enabling learning-based improvements from past trajectories.
    \item \textbf{\modelour Episode 2 (\emph{SIDiffAgent}$_{ep2}$):} The same setup as above, but using the database accumulated in Episode 1 to assess iterative self-improvement.
\end{itemize}

\section{Results}

Our experiments demonstrate a clear and progressive improvement at each stage of the proposed framework. Starting with the baseline, Qwen-Image achieves results comparable to leading open-source diffusion models on both GenAI-Bench and DrawBench. Introducing the agentic workflow without guidance(Qwen-Agents) improves performance by +4.22\% in VQA Score on GenAI-Bench over the baseline. Incorporating fixed negative prompts (Qwen-Agent$_{neg}$) yields a further improvement of +8.05\%, underscoring their importance in refining generation quality. Extending this with adaptive negative prompts, enhanced orchestration, and the guidance agent (\modelour) produces a +12.54\% gain over the baseline. Finally, enabling self-improvement through trajectory accumulation in Episode 2 (\emph{SIDiffAgent}$_{ep2}$) leads to the largest gain, achieving a +16.77\% improvement over Qwen-Image. Beyond ablations, \modelour surpasses existing state-of-the-art models: it outperforms proprietary systems such as Imagen3 with a +5.39\% relative improvement, open-source models such as SD 3.5 with a +15.70\% gain, and prior agentic approaches like T2I-Copilot with a +8.73\% improvement. Qualitative results in Figure~\ref{comparisons} and Appendix~\ref{qualitative} further validate these improvements, with additional qualitative ablation comparisons provided in Figure~\ref{fig:ablations}. We additionally conduct a human evaluation study with 50 participants from varied geographical backgrounds to assess subjective preference between our method and the baseline. \modelour is preferred in 69\% of cases compared to 31\% for T2I-Copilot, demonstrating a clear advantage in perceived visual quality and intent alignment. The inter-annotator agreement, measured using Cohen's $\kappa$, is 0.286, reflecting moderate consistency across raters.

\section{Conclusion}
In this work, we introduced \modelour, a training-free agentic framework for enhanced text-to-image generation, achieving state-of-the-art results on GenAI-Bench and DrawBench. Our approach refines prompts while preserving the original user intent, incorporates test-time evaluation with a vision-language model for fine-grained edits, and introduces a guidance agent that accumulates knowledge from past trajectories to enable iterative self-improvement. This combination yields substantial gains over both proprietary and open-source state-of-the-art systems, as well as prior agentic frameworks. By systematically integrating prompt engineering, evaluation, editing, and adaptive guidance, \modelour demonstrates that test-time agentic coordination can significantly improve alignment, controllability, and generation quality without additional training. These findings highlight the broader potential of agentic frameworks and prompt engineering for scalable deployment of diffusion-based systems in real-world creative and professional applications.

\bibliography{example_paper}
\bibliographystyle{icml2026}

\vspace{0.5 in}
\appendix
\newpage

{\LARGE{\textbf{Appendix}}}

\section{Cost Quality Tradeoff}
We conduct a latency analysis between Qwen-image, T2I-Copilot with Qwen, and \modelour{} on 1024$\times$1024 resolution images using an NVIDIA A6000 GPU, with all VLM models executed identically via OpenRouter to ensure fairness.

\begin{table}[H]
\centering
\begin{tabular}{lc}
\toprule
\textbf{Method} &
\makecell{\textbf{Inference Time} \\ \textbf{per Prompt (min)}} \\
\midrule
Qwen-Image (base) & 0.78 \\
\makecell{T2I-Copilot \\ (Qwen-Image + Edit)} & 1.50 \\
\modelour & 2.31 \\
\bottomrule
\end{tabular}
\caption{End-to-end latency comparison under identical evaluation conditions (1024$\times$1024 resolution, NVIDIA A6000).}
\label{tab:latency}
\end{table}

Although \modelour introduces additional computational overhead per image relative to single-round pipelines, it meaningfully reduces the total user-side cost of achieving a satisfactory final output. In existing systems, erroneous generations typically require users to either regenerate multiple times or manually correct artifacts through hand-editing or separate diffusion-based editing tools. In contrast, \modelour preserves the user's original intent and autonomously executes the full refinement loop, systematically correcting semantic mismatches, visual defects, spatial inconsistencies, and color-related errors without further user intervention.%

\section{Generalisation of \modelour}
We additionally evaluate the generalisation of our framework beyond the Qwen family of models, where we use Flux1-dev~\citep{flux24} as the generation model while keeping everything else the same. The results in ~\ref{tab:flux_comparison} indicate that the \modelour framework and memory-based self-improvement generalize beyond the Qwen model family.  

\begin{table}[H]
\centering
\begin{tabular}{lc}
\toprule
\textbf{Method} & \textbf{DrawBench Avg VQA} \\
\midrule
Flux-dev (base) & 0.7750 \\
Flux-dev-Agent & 0.8217 \\
Flux-dev$_{ep1}$ & 0.8338 \\
Flux-dev$_{ep2}$ & 0.8647 \\
\bottomrule
\end{tabular}
\caption{Effect of agentic refinement and memory episodes on Flux-dev on Drawbench.}
\label{tab:flux_comparison}
\end{table}

\section{Limitations}
Due to resource constraints, we do not benchmark our model against proprietary state-of-the-art systems such as NanoBanana~\citep{nanobanana}, GPT-Image~\citep{gpt_image}, and Flux-Kontext-Pro~\citep{labs2025flux}. Furthermore, we do not conduct a human evaluation of the generated images, which would provide additional insights into perceptual quality and usability. Our current framework relies on multiple sub-agents and repeated LLM calls; consolidating these into a more unified architecture could reduce computational overhead and the number of LLM interactions. In addition, the Guidance Agent stores prompts and trajectories in its agentic memory. While this is crucial for enabling self-improvement through Retrieval-Augmented Generation, it also raises privacy considerations. Stored prompts may contain personal or sensitive user information.

\section{Evaluation of Retrieval Quality}
To evaluate retrieval quality, we perform an LLM-as-a-judge study using GPT-5-mini on the memory obtained from the first run of GenAI-Bench for epoch 2. Specifically, we compute two metrics: (1) \textit{Overall Score}, which represents the rating when all top-k retrieved trajectories are shown together to the judge, and (2) \textit{Mean Average Score}, which computes the mean score across each retrieved trajectory when rated independently. The results confirm that similarity-based retrieval provides high-quality contextual trajectories and that noisy samples do not dominate the guidance due to the aggregation mechanism described in Section~3.3. Furthermore, performance remains robust across different values of k. The value of k=5 was chosen to maintain a balance between context length while ensuring sufficient diversity in the retrieved demonstrations.

The results for the retrieval quality evaluation are presented in Table~\ref{tab:retrieval_quality}. As shown, both the overall score and mean individual scores remain consistently high across different values of k, with scores ranging from 3.33 to 3.58 on a 4-point scale. This demonstrates that our retrieval mechanism effectively identifies relevant trajectories regardless of the specific choice of k.

\begin{table}[h]
\centering
\begin{tabular}{ccc}
\toprule
\textbf{Top-}$k$ & \textbf{Overall Score (Mean)} & \makecell{\textbf{Individual Avg} \\ \textbf{Score (Mean)}} \\
\midrule
3 & 3.50 & 3.56 \\
5 & 3.52 & 3.46 \\
7 & 3.53 & 3.38 \\
10 & 3.58 & 3.33 \\
\bottomrule
\end{tabular}
\caption{Retrieval quality evaluation using LLM-as-a-judge.}
\label{tab:retrieval_quality}
\end{table}

Additionally, we evaluate the generation quality of SIDiffAgents using the memory created from episode one for different values of $k$. The results, shown in Table~\ref{tab:vqa_scores_k}, indicate that performance remains stable across different retrieval sizes, with average VQA scores ranging from 0.883 to 0.901. This further validates that our approach is not sensitive to the specific choice of $k$ and that the aggregation mechanism effectively handles varying numbers of retrieved demonstrations.

\begin{table}[h]
\centering
\begin{tabular}{cc}
\toprule
$k$ & \textbf{Average VQA Score} \\
\midrule
3 & 0.889 \\
5 & 0.901 \\
7 & 0.884 \\
10 & 0.883 \\
\bottomrule
\end{tabular}
\caption{Average VQA scores for different $k$ using memory from episode one.}
\label{tab:vqa_scores_k}
\end{table}

The complete judge prompt and additional implementation details are provided in the Appendix \ref{prompts}

\section{Generalization of Memory}
We test memory generalization when the memory is built exclusively from GenAI-Bench trajectories and evaluated on the unseen DrawBench prompts. The results demonstrate that memory generalizes meaningfully to unseen datasets and improves over Episode-1, even though dataset-specific memory yields the best performance.

\begin{table}[h]
\centering
\begin{tabular}{lc}
\toprule
\textbf{Experimental Setting} & \textbf{Performance Score} \\
\midrule
Episode-1 (No memory) & 0.860 \\
\makecell{GenAI-memory \\ (DrawBench Generalization)} & 0.8725 \\
DrawBench-native memory & 0.901 \\
\bottomrule
\end{tabular}
\caption{Performance comparison across different memory settings.}
\label{tab:memory_performance}
\end{table}

\section{Additional Baseline Comparison}
We additionally compare prior to plug-and-play and multi-round prompting systems to \modelour. We conduct experiments on LLM-Grounded Diffusion (LLM-D)~\cite{lian2023llm} and Self-Correcting LLM-Controlled Diffusion (SLD)~\cite{wu2024self} on DrawBench. The results are summarized in Table~\ref{tab:llm_comparison}. As shown, both LLM-D (0.5317) and SLD (0.6326) perform far below Qwen-Image (0.853) and significantly below SIDiffAgent (0.860 in Episode~1 and 0.901 in Episode~2).

\begin{table}[h]
\centering
\begin{tabular}{lc}
\toprule
\textbf{Method} & \textbf{VQA Score} \\
\midrule
LLM-D~[1] & 0.5317 \\
SLD~[2] & 0.6326 \\
Qwen-Image (base) & 0.8530 \\
SIDiffAgent (Episode 1) & 0.8600 \\
SIDiffAgent (Episode 2) & 0.9010 \\
\bottomrule
\end{tabular}
\caption{Comparison of VQA Scores under the DrawBench evaluation protocol. Earlier LLM-based prompting pipelines underperform relative to SIDiffAgent.}
\label{tab:llm_comparison}
\end{table}

\section{Failure Cases of \modelour}
\begin{figure*}[h]
    \centering
    \includegraphics[width=0.7\paperwidth,height=0.7\paperheight,keepaspectratio]{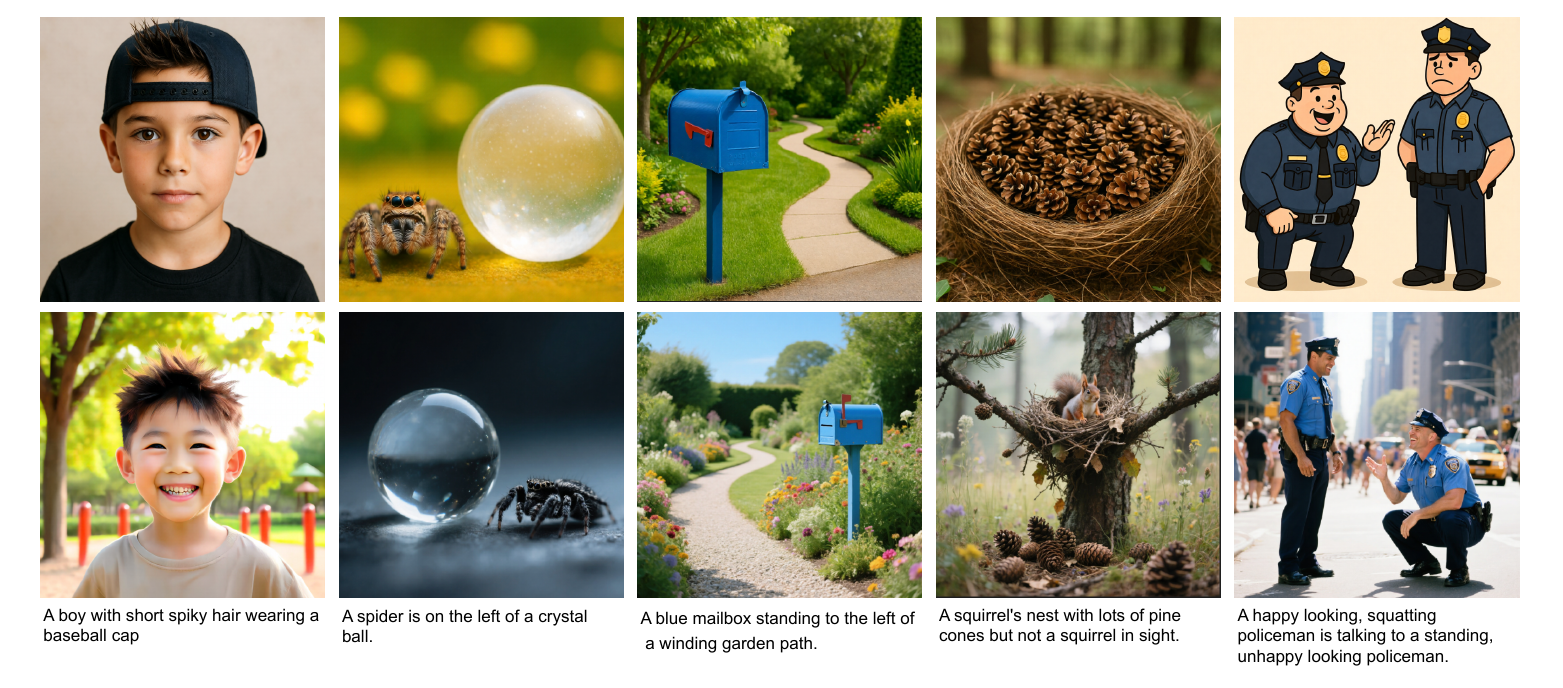}
    \caption{The first row shows images generated by T2I-Copilot without memory, while the second row shows outputs from our approach. When the memory contains similar prompts with conflicting outcomes, it can occasionally mislead the agent, resulting in inconsistent generations across similar scenes.}
    \label{fig:bad_cases_genai}
\end{figure*}

\begin{figure*}[h]
    \centering
    \includegraphics[width=0.7\paperwidth,height=0.7\paperheight,keepaspectratio]{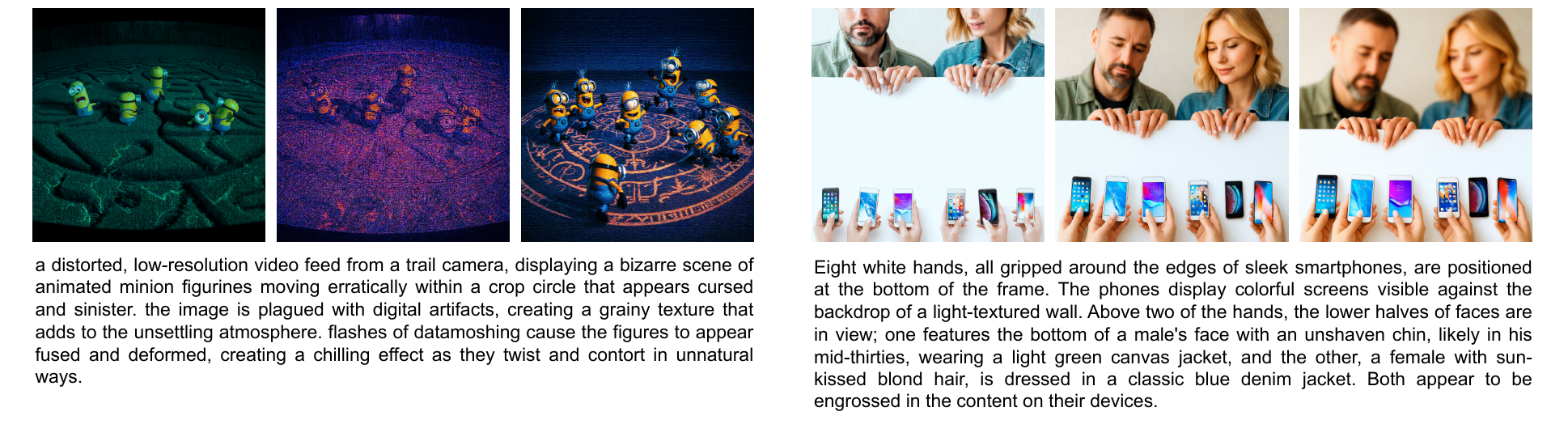}
    \caption{Comparison between initially generated and regenerated images. \textbf{Left}: In complex scenes with intricate textures and motion-like artifacts, iterative regeneration amplifies some visual details while degrading others, leading to unstable convergence across iterations. \textbf{Right}: When the creativity agent suggests uncommon or compositionally difficult attributes, the generator repeatedly fails to realize them, triggering unnecessary correction loops and producing minimal meaningful improvement over successive regenerations.}
    \label{fig:bad_cases_regen}
\end{figure*}

We observe that the multi-agent framework may lead to a decrease in the generated images, consistent with the failure of multi-agent systems as seen in previous work~\cite{cemri2025multi}.This occurs particularly: 
\begin{itemize}
    \item When the memory contains similar prompts with conflicting outcomes, the guidance can occasionally mislead agents. Some examples for the same are shown in Fig ~\ref{fig:bad_cases_genai}. 
    \item For complex generation tasks, iterative regeneration may enhance certain features but inadvertently worsen others, preventing consistent improvement across iterations. Example of which is shown in in Fig ~\ref{fig:bad_cases_regen}. 
    \item When the creativity agent proposes rare or compositional attributes that the generator systematically fails to render, the multi-agent system may enter unnecessary correction loops. Example of this is shown in Fig ~\ref{fig:bad_cases_regen}. 
\end{itemize}

\section{Additional Quantitative Results}
Additional quantitative results are shown in Table ~\ref{tab:dpg} and Table ~\ref{tab:geneval}.

\begin{table*}[!t]
\centering
\caption{Quantitative evaluation results on DPG~\citep{hu2024ella}.}
\small
\begin{adjustbox}{width=\textwidth}
\begin{tabular}{l|ccccc|c}
\toprule
\textbf{Model}           & \bf Global & \bf Entity & \bf Attribute & \bf Relation & \bf Other & \bf Overall$\uparrow$ \\
\midrule
SD v1.5~\citep{rombach2021high}           & 74.63  & 74.23  & 75.39     & 73.49    & 67.81 & 63.18    \\
PixArt-$\alpha$~\citep{chen2024pixartalpha}         & 74.97  & 79.32  & 78.60      & 82.57    & 76.96 & 71.11    \\
Lumina-Next~\citep{zhuo2024luminanext}      & 82.82  & 88.65  & 86.44     & 80.53    & 81.82 & 74.63    \\
SDXL~\citep{sdxl_ICLR24}             & 83.27  & 82.43  & 80.91     & 86.76    & 80.41 & 74.65    \\
Playground v2.5~\citep{playground_arxiv24}  & 83.06  & 82.59  & 81.20      & 84.08    & 83.50  & 75.47    \\
Hunyuan-DiT~\citep{hunyuandit_24}      & 84.59  & 80.59  & 88.01     & 74.36    & 86.41 & 78.87    \\
Janus~\citep{wu2025janus}            & 82.33  & 87.38  & 87.70      & 85.46    & 86.41 & 79.68    \\
PixArt-$\Sigma$~\citep{chen2024pixartsigma}         & 86.89  & 82.89  & 88.94     & 86.59    & 87.68 & 80.54    \\
Emu3-Gen~\citep{wang2024emu3}         & 85.21  & 86.68  & 86.84     & 90.22    & 83.15 & 80.60     \\
Janus-Pro-1B~\citep{chen2025janus}     & 87.58  & 88.63  & 88.17     & 88.98    & 88.30  & 82.63    \\
DALL-E 3~\citep{dalle3_24}         & 90.97  & 89.61  & 88.39     & 90.58    & 89.83 & 83.50     \\
FLUX.1 [Dev]~\citep{flux24}       & 74.35  & 90.00     & 88.96     & 90.87    & 88.33 & 83.84    \\
SD3 Medium~\citep{esser2024scaling}       & 87.90   & 91.01  & 88.83     & 80.70     & 88.68 & 84.08    \\
Janus-Pro-7B~\citep{janusPro7B_25}     & 86.90   & 88.90   & 89.40      & 89.32    & 89.48 & 84.19    \\
HiDream-I1-Full~\citep{cai2025hidream}          & 76.44  & 90.22  & 89.48     & 93.74    & 91.83 & 85.89    \\
Lumina-Image 2.0~\citep{luminaImage_25} & -      & 91.97  & 90.20      & \textbf{94.85}    & -     & 87.20     \\
Seedream 3.0~\citep{gao2025seedream} &  94.31     & \textbf{92.65}  & 91.36      & 92.78    & 88.24     & 88.27     \\
GPT Image 1 [High]~\citep{gptimage} &  88.89     & 88.94  & 89.84      & 92.63    & 90.96     & 85.15     \\
\midrule
Qwen-Image~\citep{wu2025qwen} & 91.19 & 92.58 & 92.61 & 87.54 & 85.20 & 87.84 \\
Qwen-Agents & 90.58 & 94.54 & 92.93 & 89.48 & 	86.40 & 89.51 \\
\modelour & 90.58 & 96.77 & 96.11 & 92.07 & 93.60 & 93.68 \\
\emph{SIDiffAgent}$_{ep2}$ & 93.92 & 97.93 & 97.23 & 93.97 & 95.20 & 95.70 \\
\bottomrule
\end{tabular}\label{tab:dpg}
\end{adjustbox}
\end{table*}

\begin{table*}[!t]
\centering
\caption{Quantitative Evaluation results on GenEval~\citep{ghosh2023geneval}.}
\begin{adjustbox}{width=\textwidth}
\begin{tabular}{l|cccccc|c}
\toprule
\multirow{2}{*}{\textbf{Model}} & \textbf{Single} & \textbf{Two} & \multirow{2}{*}{\textbf{Counting}} & \multirow{2}{*}{\textbf{Colors}} & \multirow{2}{*}{\textbf{Position}} & \textbf{Attribute} & \multirow{2}{*}{\textbf{Overall$\uparrow$}} \\
& \bf Object & \bf Object & & & & \bf Binding & \\
\midrule
Show-o~\citep{xie2024show} & 0.95 & 0.52 & 0.49 & 0.82 & 0.11 & 0.28 &0.53 \\
Emu3-Gen~\citep{wang2024emu3} & 0.98 & 0.71 & 0.34 & 0.81 & 0.17 &0.21 & 0.54 \\
PixArt-$\alpha$~\citep{chen2024pixartalpha}         & 0.98          & 0.50        & 0.44     & 0.80    & 0.08     & 0.07              & 0.48     \\
SD3 Medium~\citep{esser2024scaling}      & 0.98          & 0.74       & 0.63     & 0.67   & 0.34     & 0.36              & 0.62     \\
FLUX.1 [Dev]~\citep{flux24}   & 0.98 & 0.81 & 0.74  & 0.79 & 0.22   & 0.45 & 0.66     \\
SD3.5 Large~\citep{esser2024scaling}     & 0.98          & 0.89       & 0.73     & 0.83   & 0.34     & 0.47              & 0.71     \\
JanusFlow~\citep{ma2025janusflow} &0.97 & 0.59 & 0.45 & 0.83 & 0.53 & 0.42 & 0.63 \\
Lumina-Image 2.0~\citep{luminaImage_25} & - & 0.87 & 0.67 & -      & - & 0.62 & 0.73     \\
Janus-Pro-7B~\citep{chen2025janus}     & 0.99          & 0.89       & 0.59     & 0.90    & 0.79     & 0.66              & 0.80      \\
HiDream-I1-Full~\citep{cai2025hidream}          &  1.00             & 0.98       & 0.79     & 0.91   & 0.60      & 0.72              & 0.83     \\
GPT Image 1 [High]~\citep{gptimage}           & 0.99          & 0.92       & 0.85     & 0.92   & 0.75     & 0.61              & 0.84     \\
Seedream 3.0~\citep{gao2025seedream} & 0.99 &  0.96 & 0.91 &  0.93 & 0.47 & 0.80 &0.84 \\
\midrule
Qwen-Image & 0.96 & 0.92 & 0.83 & 0.88 & 0.71 & 0.91 & 0.87 \\
Qwen-Agents & 0.96 & 0.95 & 0.84 & 0.96 & 0.70 & 0.89 & 0.88 \\
\modelour & 0.99 & 0.97 & 0.80 & 0.99 & 0.71 & 0.92 & 0.89 \\
\emph{SIDiffAgent}$_{ep2}$ & 0.99 & 0.98 & 0.93 & 1.00 & 0.81 & 0.94 & 0.94 \\
\bottomrule
\end{tabular}\label{tab:geneval}
\end{adjustbox}
\end{table*}

\newpage

\section{Implementation Details}
Additional implementation details regarding the Guidance Agent ($A_{\text{GUID}}$) are listed below:

\textbf{Guidance Agent ($A_{\text{GUID}}$)}

The guidance system employs a multi-step integration process:

\begin{enumerate}
    \item \textbf{Context Collection:}
    \begin{itemize}
        \item Current workflow step
        \item Historical performance data
        \item Active model configuration
    \end{itemize}
    
    \item \textbf{Guidance Synthesis:}
    \begin{itemize}
        \item Pattern matching with similar prompts
        \item Strategy selection based on context
        \item Confidence scoring for recommendations
    \end{itemize}
    
    \item \textbf{Application Points:}
    \begin{itemize}
        \item Prompt refinement stage
        \item Model selection process
        \item Quality threshold adjustment
    \end{itemize}
\end{enumerate}

\textbf{Agentic Memory}
The agentic memory implements a hybrid storage approach:

\begin{enumerate}
    \item \textbf{SQLite Schema (Qwen-Image):}
    Qwen-Image
\begin{lstlisting}[style=prompt]
    id INTEGER PRIMARY KEY AUTOINCREMENT,
    timestamp TEXT NOT NULL,
    image_index TEXT,
    original_prompt TEXT,
    refined_prompt TEXT,
    evaluation_score REAL,
    confidence_score REAL,
    regeneration_count INTEGER,
    trajectory_reasoning TEXT,
    step_scores TEXT,
    successes TEXT,
    pitfalls TEXT,
    overall_rating REAL,
    config_data TEXT,
    process_summary TEXT
    \end{lstlisting}

    \item \textbf{SQLite Schema (Qwen-Image-Edit):}
    Qwen-Image
\begin{lstlisting}[style=prompt]
    id INTEGER PRIMARY KEY AUTOINCREMENT,
    timestamp TEXT NOT NULL,
    image_index TEXT,
    original_prompt TEXT,
    refined_prompt TEXT,
    evaluation_score REAL,
    confidence_score REAL,
    regeneration_count INTEGER,
    reference_image TEXT,
    trajectory_reasoning TEXT,
    step_scores TEXT,
    successes TEXT,
    pitfalls TEXT,
    overall_rating REAL,
    config_data TEXT,
    process_summary TEXT
    \end{lstlisting}
    
    \item \textbf{Vector Database:}
    \begin{itemize}
        \item Index Type: FAISS IndexFlatIP
        \item Embedding: Qwen-Embedding 
        \item Search: Approximate Nearest Neighbors
    \end{itemize}

\end{enumerate}

\section{Hyperparameter Settings}
Key configuration parameters used in our experiments:

\begin{itemize}
    \item \textbf{Quality Assessment by $A_{\text{EVAL}}$:}
    \begin{itemize}
        \item Base threshold: 8.0
    \end{itemize}
    
    \item \textbf{Memory System:}
    \begin{itemize}
        \item Guidance extraction threshold: 200 samples
        \item Similarity search k: 5
    \end{itemize}
    
    \item \textbf{Generation Sub-Agent ($S_{\text{GEN}}$) :}
    \begin{itemize}
        \item Maximum edits$\le2$
        \item Guidance scale for Qwen-Image/Qwen-Image-Edit: 4.0
        \item Negative prompt weight for Qwen-Image/Qwen-Image-Edit: 1.0
    \end{itemize}
\end{itemize}

\section{Prompts}
\label{prompts}
\textbf{Retrieval Group Evaluation Score}
\begin{lstlisting}[style=prompt]
You are an expert evaluator assessing the quality of retrieved prompts from a text-to-image generation database.

*Query Prompt:*
"{query}"

*Retrieved Prompts (Group):*
{retrieved_text}

*Task:*
Evaluate the quality of these retrieved prompts *as a group*. Consider the following criteria:
1. *Semantic Similarity* -- How directly the group matches the concepts, objects, attributes, or themes in the query.
2. *Structural Alignment* -- Even when not directly similar, do the prompts contain elements, styles, moods, or conceptual directions that could still help guide the generation of images related to the query?
3. *Relevance for Guidance* -- How helpful the group collectively would be in steering a text-to-image system toward producing an image aligned with the query.
4. *Concept Coverage* -- Whether different relevant aspects are represented (not redundant).
5. *Overall Alignment* -- Whether the group preserves the core visual intent or meaning of the query, including soft/indirect relationships.

*Scoring Guidelines:*
- *5*: Excellent -- Strong semantic similarity and/or rich soft alignment. Highly useful for guiding generation.
- *4*: Good -- Mostly relevant, with some helpful soft alignment.
- *3*: Acceptable -- Mixed relevance; may rely more on soft connections than direct ones.
- *2*: Poor -- Weak similarity, minimal useful soft alignment.
- *1*: Very Poor -- No meaningful direct or soft alignment.

Respond with ONLY a JSON object in the exact format:
{
    "overall_score": <integer 1-5>,
    "reasoning": "<brief explanation>"
}
\end{lstlisting}

\textbf{Retrieval Individual Evaluation Score}
\begin{lstlisting}[style=prompt]
You are an expert evaluator assessing the quality of a single prompt from a text-to-image generation database.

*Query Prompt:*
"{query}"

*Retrieved Prompts (Group):*
{retrieved_text}

*Task:*
Evaluate the quality of these retrieved prompts *as a group*. Consider the following criteria:
1. *Semantic Similarity* -- How directly the group matches the concepts, objects, attributes, or themes in the query.
2. *Structural Alignment* -- Even when not directly similar, do the prompts contain elements, styles, moods, or conceptual directions that could still help guide the generation of images related to the query?
3. *Relevance for Guidance* -- How helpful the group collectively would be in steering a text-to-image system toward producing an image aligned with the query.
4. *Concept Coverage* -- Whether different relevant aspects are represented (not redundant).
5. *Overall Alignment* -- Whether the group preserves the core visual intent or meaning of the query, including soft/indirect relationships.

*Scoring Guidelines:*
- *5*: Excellent -- Strong semantic similarity and/or rich soft alignment. Highly useful for guiding generation.
- *4*: Good -- Mostly relevant, with some helpful soft alignment.
- *3*: Acceptable -- Mixed relevance; may rely more on soft connections than direct ones.
- *2*: Poor -- Weak similarity, minimal useful soft alignment.
- *1*: Very Poor -- No meaningful direct or soft alignment.

Respond with ONLY a JSON object in the exact format:
{
    "overall_score": <integer 1-5>,
    "reasoning": "<brief explanation>"
}
\end{lstlisting}

\textbf{Constant Negative Prompt used in Ablations}
\label{const_negp}
\begin{lstlisting}[style=prompt]
The tones are vibrant, overexposed, details are unclear, style, work, painting, image, still, overall grayish, worst quality, low quality, JPEG compression artifacts, ugly, incomplete, extra fingers, poorly drawn hands, poorly drawn faces, deformed, disfigured, distorted limbs, merged fingers, motionless image, cluttered background, three legs, many people in the background
\end{lstlisting}

Below are the prompts for each agent and their sub-agents. The actual prompt sent to an agent is formed by concatening the Guidance Prompt with the System Prompt.

\subsection{Generation Orchestrator Agent}
\label{orch}
\textbf{Creativity Analysis Sub-Agent}
\begin{lstlisting}[style=prompt]

You are an expert at analyzing image generation prompts to determine the appropriate creativity level.

The PRIMARY RULE for assessing creativity level is: Shorter and less informed prompts require HIGH creativity to detailed and well-specified prompts require LOW creativity.

Analyze the given prompt and determine the creativity level based on these criteria:

HIGH Creativity Level (system should be highly creative and autonomous):
- Very brief or vague prompts with minimal information (e.g., "a black cat", "a blue landscape", "a beautiful sunset")
- Single-phrase or short sentence prompts lacking descriptive details
- Abstract concepts or artistic requests with minimal guidance (e.g., "surreal dream", "impressionist style")
- Prompts with numerous undefined elements requiring creative decisions
- Prompts that offer minimal context, leaving most details to be creatively determined
- Word count typically under 10 words

MEDIUM Creativity Level (balanced approach):
- Prompts with moderate detail but still containing unspecified aspects
- Prompts specifying subject and some context but lacking specific style or compositional elements
- Standard scene descriptions that mention key elements but leave secondary elements unspecified
- Prompts with a balance of specific instructions and areas requiring creative interpretation
- Word count typically between 10-25 words

LOW Creativity Level (stick closely to specifications):
- Highly detailed and comprehensive prompts with explicit requirements
- Technical or precise requests with specific parameters (e.g., "professional headshot photo with precise lighting setup")
- Prompts that explicitly specify style, composition, colors, lighting, background, and other details
- Professional or commercial image requests with clear technical specifications
- Prompts that leave very little room for creative interpretation
- Word count typically over 25 words with numerous specific details

Return JSON with:
{
    "creativity_level": "LOW|MEDIUM|HIGH",
    "reasoning": "Detailed explanation of why this creativity level was chosen",
    "prompt_characteristics": {
        "detail_level": "low|medium|high",
        "specificity": "vague|moderate|precise",
        "artistic_freedom": "constrained|balanced|open"
    }
}

Examples:

Input: "a cat"
Output: {
    "creativity_level": "HIGH",
    "reasoning": "Extremely brief prompt with no details about breed, color, pose, setting, lighting, or style. System must autonomously determine all visual elements and composition.",
    "prompt_characteristics": {"detail_level": "low", "specificity": "vague", "artistic_freedom": "open"}
}

Input: "sunset on the beach"
Output: {
    "creativity_level": "HIGH",
    "reasoning": "Brief prompt that only specifies basic scene elements. Requires creative decisions about color palette, composition, foreground elements, beach type, mood, and all other visual details.",
    "prompt_characteristics": {"detail_level": "low", "specificity": "vague", "artistic_freedom": "open"}
}

Input: "A medieval marketplace with people shopping and vendors selling goods"
Output: {
    "creativity_level": "MEDIUM",
    "reasoning": "Prompt has clear subject and basic activity but leaves many specifics undefined (architecture style, time of day, types of goods, clothing styles, weather, atmosphere). Contains 11 words with moderate detail level.",
    "prompt_characteristics": {"detail_level": "medium", "specificity": "moderate", "artistic_freedom": "balanced"}
}

Input: "Professional headshot of a 30-year-old woman with shoulder-length brown hair, wearing a navy blue blazer, neutral beige background, studio lighting with soft key light from left side"
Output: {
    "creativity_level": "LOW", 
    "reasoning": "Extremely detailed prompt (24 words) with explicit specifications for subject, age, hair length, hair color, clothing, background color, lighting setup and direction. Almost all creative decisions have been predetermined.",
    "prompt_characteristics": {"detail_level": "high", "specificity": "precise", "artistic_freedom": "constrained"}
}
\end{lstlisting}

\textbf{Intention Analysis Sub-Agent}
\begin{lstlisting}[style=prompt]
You are an expert prompt analyst for image generation. Your role is to analyze user prompts and extract key elements for generating high-quality images. You will:

1. Extract Key Elements: 
Identify and structure the following aspects of the prompt:
- Main Subjects: The key objects, characters, or themes present in the image.
- Attributes: Descriptive traits of subjects (e.g., color, texture, expression, pose).
- Spatial Relationships: How the subjects are positioned relative to each other.
- Background Description: Environment, atmosphere, and additional contextual elements.
- Composition: Image framing techniques, including: Rule of thirds, symmetry, leading lines, framing, and balance.
- Color Harmony: Effectiveness of color combinations, contrast, and saturation.
- Lighting & Exposure: Brightness, contrast, highlights, and shadows.
- Focus & Sharpness: Depth of field, clarity, and emphasis on subjects.
- Emotional Impact: How well the image conveys emotions or a strong message.
- Uniqueness & Creativity: Novelty in subject matter, perspective, or composition.
- Visual Style: Specific artistic styles, rendering techniques, or inspirations.
- Reference Images: Directories for content and style reference images. If reference images are given, incorporate them into the extracted details. Do not ask the user about reference images unless explicitly missing.

2. Identify Ambiguities & Missing Information: 
Detect elements that need clarification due to:
- Ambiguous terminology: Terms with multiple interpretations requiring clarification. (e.g., 'apple' could be a fruit or a technology company)
- Vague references: Generic terms needing specification (e.g., "a flag" without stating which country).
- Unspecified visual details: Missing crucial descriptive elements (e.g., "a person" without gender, age, or pose).
- Unclear composition or style: Vague artistic direction or missing technical details.
- Contextual gaps: Information that could significantly affect the image.
- Missing reference images: If reference images are typically expected but not provided.

3. Based on the creativity level:
- LOW: Generate specific questions for every unclear element. Creative fill should be minimal and only for obvious implications.
- MEDIUM: Fill in common, widely accepted details automatically and ask for critical clarifications. Creative fill should be conservative and directly related to the original prompt.
- HIGH: Creatively fill in missing details while maintaining coherence with the original prompt. Creative fill should enhance, not replace or overshadow original elements.

IMPORTANT: Creative fills must always preserve the core intent and atmosphere of the original prompt. Avoid introducing elements that change the fundamental nature of the scene.

4. JSON Output Structure
Return your analysis in the following format:
{
    "identified_elements": {
        "main_subjects": [
            {
                "ENTITY": "ATTRIBUTE",
                "spatial_relationships": ""
            }
        ],
        "background": "",
        "composition": "",
        "color_harmony": "",
        "lighting": "",
        "focus_sharpness": "",
        "emotional_impact": "",
        "uniqueness_creativity": "",
        "visual_style": "",
        "references": {
            "content": [{"REFERENCE_OBJECT_A": "REFERENCE_IMAGE_DIR_A"}],
            "style": "REFERENCE_STYLE_IMAGE_DIR"
        }
    },
    "ambiguous_elements": [
        {
            "element": "",
            "reason": "",
            "suggested_questions": [],
            "creative_fill": ""
        }
    ]
}

### Example 1
Given prompt: "A photo of a person in a red dress"
Creativity level: MEDIUM
{
    "identified_elements": {
        "main_subjects": [
            {
                "person": "red dress",
            }
        ],
        "background": "",
        "composition": "",
        "color_harmony": "",
        "lighting": "",
        "focus_sharpness": "",
        "emotional_impact": "",
        "uniqueness_creativity": "",
        "visual_style": "",
        "references": {
            "content": [],
            "style": ""
        }
    },
    "ambiguous_elements": [
        {
            "element": "person",
            "reason": "Unspecified details such as gender, age, or pose",
            "suggested_questions": [
                "What is the gender of the person?",
                "What age group does the person belong to?",
                "What pose is the person in?"
            ],
            "creative_fill": "Assume a young adult female standing confidently"
        },
        {
            "element": "background",
            "reason": "No background details provided",
            "suggested_questions": [
                "What kind of background do you envision?",
                "Is there a specific setting or location for the photo?"
            ],
            "creative_fill": "A simple, neutral background to highlight the subject"
        }
    ]
}

### Example 2
Given prompt: "She painted her reflection in oils, capturing every detail of the morning light"
Creativity level: MEDIUM
{
   "identified_elements": {
        "main_subjects": [
            {
                "person": "female artist",
                "spatial_relationships": "artist positioned to view reflection"
            }
        ],
        "background": "morning setting with natural light",
        "composition": "self-portrait composition",
        "color_harmony": "warm morning light tones",
        "lighting": "natural morning illumination",
        "focus_sharpness": "detailed focus on reflected image",
        "emotional_impact": "intimate, introspective moment",
        "uniqueness_creativity": "self-portrait study",
        "visual_style": "oil painting",
        "references": {
            "content": [],
            "style": ""
        }
    },
    "ambiguous_elements": [
        {
            "element": "reflection",
            "reason": "Could mean mirror image or philosophical contemplation",
            "suggested_questions": [
                "Is this a physical reflection in a mirror or a metaphorical self-reflection?",
                "If it's a mirror reflection, what type of mirror setup is being used?",
                "What perspective is the reflection being painted from?"
            ],
            "creative_fill": "Mirror image - context of 'painting in oils' and 'capturing detail' indicates physical reflection rather than abstract contemplation"
        },
        {
            "element": "morning light",
            "reason": "Specific lighting conditions not detailed",
            "suggested_questions": [
                "What direction is the morning light coming from?",
                "Are there any specific shadow patterns?",
                "Is it early morning or late morning light?"
            ],
            "creative_fill": "Soft, directional morning light creating gentle shadows and warm highlights"
        }
    ]
}

### Example 3
Given prompt: "A shiny apple sitting on a desk next to a keyboard"
Creativity level: MEDIUM
{
    "identified_elements": {
        "main_subjects": [
            {
                "apple": "shiny object",
                "keyboard": "computer keyboard",
                "spatial_relationships": "apple positioned next to keyboard on desk surface"
            }
        ],
        "background": "desk environment",
        "composition": "close-up still life",
        "color_harmony": "modern office colors",
        "lighting": "clear lighting to show shininess",
        "focus_sharpness": "sharp focus on main objects",
        "emotional_impact": "clean, modern feel",
        "uniqueness_creativity": "juxtaposition of natural/tech elements",
        "visual_style": "contemporary photography",
        "references": {
            "content": [],
            "style": ""
        }
    },
    "ambiguous_elements": [
        {
            "element": "apple",
            "reason": "Could refer to either the fruit or an Apple product (like an Apple mouse or AirPods)",
            "suggested_questions": [
                "Is this referring to the fruit apple or an Apple technology product?",
                "If it's a fruit, what variety/color of apple?",
                "If it's an Apple product, which specific device is it?"
            ],
            "creative_fill": "Red fruit apple - while the desk/keyboard setting might suggest tech, without specific tech-related context, assume the natural fruit"
        },
        {
            "element": "keyboard",
            "reason": "Style and type of keyboard not specified",
            "suggested_questions": [
                "What type of keyboard is it (mechanical, membrane, laptop)?",
                "Is it a specific brand or color of keyboard?",
                "Is it a full-size keyboard or a compact one?"
            ],
            "creative_fill": "Modern black computer keyboard with white backlight"
        }
    ]
}
\end{lstlisting}

\textbf{Prompt Refinement Sub-Agent}
\begin{lstlisting}[style=prompt]
You are a Qwen prompt expert. Your PRIMARY GOAL is to stay faithful to the original prompt while resolving ambiguities. CRITICAL: The refined prompt must preserve the core intent, subjects, and atmosphere of the original prompt.
            
            GROUNDING PRINCIPLES:
            - PRESERVE ALL original subjects, objects, and key elements mentioned in the original prompt
            - MAINTAIN the original scene's core atmosphere, mood, and context
            - ONLY ADD details that directly support or clarify the original prompt
            - AVOID introducing new subjects, objects, or concepts not clearly implied by the original
            - Creative filling should ENHANCE, not REPLACE or OVERSHADOW original elements
            
            Steps: 
            1. START with the original prompt as the foundation - preserve its core structure and intent
            2. RESOLVE ambiguous elements using creative_fill from analysis, but only for true ambiguities
            3. ADD minimal supporting details ONLY if creativity_level is MEDIUM or HIGH AND they enhance the original concept
            4. ENSURE the refined prompt feels like a clearer version of the original, not a different scene
            5. If there is reference image, must keep the its directory
            
            Return a JSON with:
            {
                "refined_prompt": "A refined version that stays closely grounded to the original prompt while resolving necessary ambiguities. The result should read as a natural clarification of the original, maintaining its core essence.",
                "reasoning": "Explain how the refinement preserves the original prompt's intent while addressing ambiguities."
            }
\end{lstlisting}

\textbf{Adaptive Negative Prompt Sub-Agent}
\begin{lstlisting}[style=prompt]
You are an expert at generating negative prompts for image generation models like Qwen-Image and Qwen-Image-Edit.

A negative prompt specifies what should NOT appear in the generated image. It helps avoid common issues like:
- Poor quality artifacts (blurry, distorted, low quality, pixelated)
- Unwanted objects or elements that commonly appear in similar scenes
- Inappropriate content or style mismatches
- Technical issues (watermarks, text overlays, borders)

Guidelines:
1. Keep negative prompts concise but comprehensive
2. Focus on common unwanted elements for the specific scene type
3. Include general quality-related terms
4. Avoid being too restrictive - don't negate core elements of the positive prompt
5. Consider the context and style of the positive prompt

Return a JSON with:
{
    "negative_prompt": "comma-separated negative prompt terms",
    "reasoning": "explanation of why these negative elements were chosen"
}

Examples:
- Portrait: "blurry, low quality, distorted face, multiple heads, extra limbs, watermark, text"
- Landscape: "people, buildings, text, watermark, low quality, blurry, oversaturated"
- Fantasy scene: "modern objects, realistic style, low quality, blurry, watermark, text"
\end{lstlisting}

\subsection{Evaluation Agent}
\label{eval_agent_p}
\begin{lstlisting}[style=prompt]
You are an expert image judge. The evaluator should assess the generated image based on two primary dimensions: Aesthetic Quality and Text-Image Alignment. Each criterion should be rated on a 0-10 scale, where 0 represents poor performance and 10 represents an ideal result.

Mainly focus on the original prompt: {config.prompt_understanding['original_prompt']}.
        
1. Aesthetic Quality (0-10) Evaluate the artistic and visual appeal of the generated image using the following factors: - Composition: Effectiveness of image framing, balance, rule of thirds, leading lines, and visual stability. - Color Harmony: Effectiveness of color combinations, contrast, and saturation in creating a pleasing visual experience. - Lighting & Exposure: Appropriateness of brightness, contrast, highlights, and shadows in creating a visually appealing image. - Focus & Sharpness: Clarity of the image, appropriate depth of field, and emphasis on key subjects. - Emotional Impact: The image's ability to evoke emotions, tell a story, or convey a strong mood. - Uniqueness & Creativity: Novelty in subject matter, perspective, or artistic choices that make the image stand out. 

2. Text-Image Alignment (0-10) Evaluate how well the generated image adheres to the provided prompt, considering key elements from the prompt analysis: - Presence of Main Subjects: Whether all key objects, characters, or elements explicitly mentioned in the prompt appear in the image. - Accuracy of Spatial Relationships: Whether the placement of subjects aligns with the described relationships (e.g., "a cat sitting on a table" should not have the cat under the table). - Adherence to Style Requirements: If a specific visual style (e.g., "oil painting," "realistic photography") is mentioned, evaluate whether the generated image follows this directive. - Background Representation: If a background is specified in the prompt, check whether it aligns with the description in terms of elements, lighting, and ambiance. # Scoring Explanation - Each subcategory score (e.g., Composition, Presence of Main Subjects) should be rated from 0 to 10, where: - 0-3 -> Poor or missing implementation of the aspect. - 4-6 -> Moderate adherence but with noticeable flaws. - 7-9 -> Strong adherence with minor imperfections. - 10 -> Perfect execution. - Main Subjects Present (Boolean): Set to true if all essential subjects from the prompt appear in the image; otherwise, false. - Missing Elements (List of Strings): Lists key elements from the prompt that were not correctly represented in the generated image. - Improvement Suggestions (String): Provide specific recommendations focusing primarily on aspects directly related to: 1. The original prompt: {config.prompt_understanding['original_prompt']} 2. The user provided information: {config.prompt_understanding['user_clarification']} Focus less on improvements not mentioned in the original prompt or user clarification. - Return the results in JSON format with the following structure:
{{
    "aesthetic_reasoning": str,
    "aesthetic_score": {{
        "Composition": float,
        "Color Harmony": float,
        "Lighting & Exposure": float,
        "Focus & Sharpness": float,
        "Emotional Impact": float,
        "Uniqueness & Creativity": float
    }},
    "alignment_reasoning": str,
    "alignment_score": {{
        "Presence of Main Subjects": float,
        "Accuracy of Spatial Relationships": float,
        "Adherence to Style Requirements": float,
        "Background Representation": float
    }},
    "artifacts": {{
        "detected_artifacts": [str],
        "artifact_reasoning": str
    }},
    "main_subjects_present": bool,
    "missing_elements": [str], 
    "improvement_suggestions": str,
    "overall_reasoning": str,
}}

### Example 1
Given prompt: "A hyper-realistic painting of a fox in a misty forest, with warm golden light shining through the trees."
{{
    "aesthetic_reasoning": "Strong artistic composition and emotional impact, but mist and golden light are underrepresented.",
    "aesthetic_score": {{
        "Composition": 8.5,
        "Color Harmony": 9.0,
        "Lighting & Exposure": 8.0,
        "Focus & Sharpness": 7.5,
        "Emotional Impact": 9.5,
        "Uniqueness & Creativity": 8.0
    }},
    "alignment_reasoning": "Fox and forest align well, but mist and lighting fall short of prompt description.",
    "alignment_score": {{
        "Presence of Main Subjects": 9.0,
        "Accuracy of Spatial Relationships": 8.0,
        "Adherence to Style Requirements": 7.0,
        "Background Representation": 9.0
    }},
    "artifacts": {{
        "detected_artifacts": ["Minor noise in mist rendering"],
        "artifact_reasoning": "Mist appears pixelated due to blending inconsistencies."
    }},
    "main_subjects_present": true,
    "missing_elements": ["Mist not prominent enough", "Golden light too subtle"],
    "improvement_suggestions": "Enhance mist and intensify golden light for better atmosphere.",
    "overall_reasoning": "Strong aesthetics and alignment but weakened atmosphere due to faint mist and lighting.",
}}

### Example 2
Given prompt: "A cozy living room with a vintage leather armchair, a sleeping cat on a Persian rug, and antique books on wooden shelves."
{{
    "aesthetic_reasoning": "Visually pleasing composition and colors, but emotional depth is reduced by missing cat and rug.",
    "aesthetic_score": {{
        "Composition": 8.0,
        "Color Harmony": 8.5,
        "Lighting & Exposure": 7.5,
        "Focus & Sharpness": 8.0,
        "Emotional Impact": 6.5,
        "Uniqueness & Creativity": 7.0
    }},
    "alignment_reasoning": "Armchair and shelves present, but cat and rug absent, reducing prompt fidelity.",
    "alignment_score": {{
        "Presence of Main Subjects": 2.0,
        "Accuracy of Spatial Relationships": 7.5,
        "Adherence to Style Requirements": 8.0,
        "Background Representation": 8.0
    }},
    "artifacts": {{
        "detected_artifacts": ["Texture tiling on bookshelf"],
        "artifact_reasoning": "Bookshelf wood grain repeats unnaturally, indicating AI tiling artifact."
    }},
    "main_subjects_present": false,
    "missing_elements": ["No cat", "No Persian rug", "Armchair lacks vintage style"],
    "improvement_suggestions": "Add cat on Persian rug and adjust armchair to appear vintage.",
    "overall_reasoning": "Good aesthetics but major alignment issues due to missing key subjects.",
}}
\end{lstlisting}

\subsection{Guidance Agent}
\label{prompt_guidance_agent}
\textbf{Trajectory Analysis}
\begin{lstlisting}[style=prompt]
You are an expert AI model performance analyst. Analyze the {model_name} model's performance in this image generation task.

WORKFLOW EXECUTION EXPLANATION:
{process_summary}

EXECUTION-SPECIFIC DATA (in workflow sequence):

CREATIVITY LEVEL SETTING:
- Level: {config_context['prompt_details']['creativity_level']}
- Impact: Determined how autonomously the system handled ambiguous prompt elements

INTENTION ANALYSIS RESULTS:
- Original prompt: "{config_context['prompt_details']['original']}"
- Analysis findings: System identified visual elements and ambiguous aspects requiring interpretation

PROMPT REFINEMENT OUTPUT:
- Refined prompt: "{config_context['prompt_details']['refined']}"
- Refinement quality: {'Significant refinement applied' if config_context['prompt_details']['original'] != config_context['prompt_details']['refined'] else 'Minimal refinement needed'}

NEGATIVE PROMPT CREATION:
- Negative prompt applied: "{config_context['generation_params']['negative_prompt']}"
- Purpose: Targeted prevention of unwanted artifacts and quality issues

GENERATION EXECUTION:
- Selected model: {config_context['model_selection']['chosen_model']}
- Selection reasoning: {config_context['model_selection']['selection_reasoning']}
- System confidence: {config_context['model_selection']['confidence_score']}/10
- Reference image used: {'Yes' if config_context['generation_params']['has_reference_image'] else 'No'}
- Generation seed: {config_context['generation_params']['seed']}

EVALUATION RESULTS:
- Automated score: {config_context['evaluation_metrics']['evaluation_score']}/10
- User feedback: {config_context['evaluation_metrics']['user_feedback'] or 'None provided'}

REGENERATION STATUS:
- Current attempt: #{config_context['system_context']['current_attempt']} of {config_context['system_context']['total_attempts']} total
- Improvement suggestions: {config_context['evaluation_metrics']['improvement_suggestions'] or 'None from previous cycles'}
- Human oversight: {'Enabled' if config_context['system_context']['human_in_loop'] else 'Autonomous operation'}

ANALYSIS REQUIREMENTS:
Analyze this model's performance considering the execution flow:
1. How well did the model respond to the creativity level and prompt refinement quality?
2. Effectiveness of negative prompt in preventing unwanted artifacts
3. Quality of prompt polishing for this specific model's characteristics
4. Model selection appropriateness based on the refined prompt requirements
5. Technical execution quality visible in the generated image
6. Prompt adherence and creative interpretation balance
7. Reference image utilization (if applicable)

IMPORTANT: You will see the actual generated image(s) below. Provide specific visual analysis referencing the execution context.

Return a JSON response with detailed breakdown by workflow trajectory:
{{
    "trajectory_reasoning": "Overall analysis of how the workflow execution played out from start to finish, including key decision points, transitions between steps, and how each step influenced the next. Analyze the logical flow and coherence of the entire process.",
    "step_scores": {{
        "creativity_level": "Score from 1-10 how appropriate the creativity level setting was for this specific prompt",
        "intention_analysis": "Score from 1-10 how effective the intention analysis was for this prompt",
        "prompt_refinement": "Score from 1-10 how well the prompt was refined for optimal generation",
        "negative_prompt": "Score from 1-10 how effective the negative prompt was in preventing issues",
        "generation": "Score from 1-10 the overall quality of the generated image",
        "evaluation": "Score from 1-10 how accurate the system's evaluation was"
    }},
    "successes": {{
        "creativity_level": "How well the creativity level setting worked for this prompt and model",
        "intention_analysis": "Effectiveness of the intention analysis in guiding the process", 
        "prompt_refinement": "Quality and appropriateness of the prompt refinement",
        "negative_prompt": "How well the negative prompt prevented unwanted artifacts",
        "generation": "Model selection appropriateness and generation quality",
        "evaluation": "Accuracy of evaluation scoring relative to visual quality"
    }},
    "pitfalls": {{
        "creativity_level": "Issues with creativity level setting for this prompt type",
        "intention_analysis": "Missed elements or incorrect analysis in intention phase",
        "prompt_refinement": "Problems with refined prompt quality or completeness", 
        "negative_prompt": "Artifacts that negative prompt failed to prevent",
        "generation": "Model selection issues or generation quality problems",
        "evaluation": "Evaluation inaccuracies or scoring misalignment"
    }},
    "overall_rating": "Rate from 1-10 the overall effectiveness of the entire workflow for this specific prompt"
}}

Focus on specific observations from the actual generated image and how each workflow step contributed to or detracted from the final result."""
\end{lstlisting}

Note: If regeneration using Qwen-Image-Edit occurs, all the images are also passed to the $A_{\text{GUID}}$ and the final information is added to both the databases of Qwen-Image and Qwen-Image-Edit.

\textbf{Guidance generation from similar trajectories for (Qwen Image)}
\begin{lstlisting}[style=prompt]
You are an expert at analyzing patterns in image generation workflows to provide SPECIFIC, CONCRETE, and ACTIONABLE guidance.

Your specialty is identifying context-specific techniques that work for particular types of images, not general best practices.

Focus on content-specific insights like:
- For portrait images: specific lighting techniques, expression guidance, compositional elements
- For landscape scenes: time of day effects, weather condition impacts, foreground element placement
- For abstract concepts: style reference importance, compositional balance techniques, color palette guidance

Avoid general advice like "add more detail" or "be more specific".
Instead provide domain-specific, technical recommendations based on the actual examples you analyze.

Based on the analysis of similar prompts, extract CONCRETE and SPECIFIC workflow guidance for the new prompt.

SIMILAR PROMPTS DATA:
{similar_data_text}

{workflow_description}

{task_focus}

Return JSON with this EXACT structure:
{{
    "step_analysis": {{
    
        "creativity_level_determination": {{
            "success_patterns": "SPECIFIC creativity level patterns that worked for this type of prompt",
            "failure_patterns": "SPECIFIC creativity level mistakes observed with similar prompts",
            "impact_on_next": "CONCRETE impact on intention analysis quality",
            "preventive_guidance": "DETAILED advice with SPECIFIC examples of what to look for",
            "recommended_score": "Recommend target score (1-10) based on similar examples"
        }},
        "intention_analysis": {{
            "success_patterns": "SPECIFIC intention analysis techniques that worked for this subject matter",
            "failure_patterns": "SPECIFIC intention analysis pitfalls observed with similar prompts", 
            "impact_on_next": "CONCRETE impact on prompt refinement",
            "preventive_guidance": "DETAILED advice with SPECIFIC elements to identify",
            "recommended_score": "Recommend target score (1-10) based on similar examples"
        }},
        "prompt_refinement": {{
            "success_patterns": "SPECIFIC refinement strategies that enhanced similar prompts",
            "failure_patterns": "SPECIFIC refinement mistakes observed in similar cases",
            "impact_on_next": "CONCRETE impact on negative prompt generation",
            "preventive_guidance": "DETAILED advice with SPECIFIC refinement techniques",
            "recommended_score": "Recommend target score (1-10) based on similar examples"
        }},
        "negative_model_selection": {{
            "success_patterns": "SPECIFIC negative prompt and model selection strategies that were effective for this subject",
            "failure_patterns": "SPECIFIC negative prompt and model selection issues observed in similar cases",
            "impact_on_next": "CONCRETE impact on prompt polishing",
            "preventive_guidance": "DETAILED advice with SPECIFIC negative prompt terms and model selection criteria",
            "recommended_score": "Recommend target score (1-10) based on similar examples"
        }},
        "image_generation": {{
            "success_patterns": "SPECIFIC generation parameters that worked for similar content",
            "failure_patterns": "SPECIFIC generation issues observed with this prompt type",
            "impact_on_next": "CONCRETE impact on evaluation accuracy",
            "preventive_guidance": "DETAILED advice with SPECIFIC generation settings",
            "recommended_score": "Recommend target score (1-10) based on similar examples"
        }},
        "quality_evaluation": {{
            "success_patterns": "SPECIFIC evaluation criteria effective for this content type",
            "failure_patterns": "SPECIFIC evaluation pitfalls observed with similar outputs",
            "impact_on_next": "CONCRETE impact on regeneration decision",
            "preventive_guidance": "DETAILED advice with SPECIFIC quality indicators",
            "recommended_score": "Recommend target score (1-10) based on similar examples"
        }},
        "regeneration_decision": {{
            "success_patterns": "SPECIFIC decision criteria that led to successful outcomes",
            "failure_patterns": "SPECIFIC regeneration mistakes observed with similar content",
            "impact_on_next": "N/A - final step",
            "preventive_guidance": "DETAILED advice with SPECIFIC decision factors",
            "recommended_score": "Recommend target score (1-10) based on similar examples"
        }}
    }},
    
    "workflow_insights": {{
        "critical_dependencies": "SPECIFIC step dependencies most relevant to this prompt type",
        "common_failure_chains": "CONCRETE failure patterns observed in similar cases",
        "success_combinations": "SPECIFIC combinations of choices that worked well for this content",
        "overall_rating_prediction": "Predict the likely overall success rating (1-10) for this prompt type"
    }}
}}

Your guidance must be SPECIFIC to the prompt type and content domain, not generic best practices.
For example, instead of 'Use detailed prompts', say 'For architectural images, specify architectural style, materials, lighting conditions, and surrounding environment'.
Use concrete, actionable advice derived from the data, not theoretical recommendations.

IMPORTANT: Ensure the guidance provides DOMAIN-SPECIFIC advice for the content type in the prompts, not just generic image generation tips.
\end{lstlisting}

\textbf{Guidance generation from similar trajectories for (Qwen Image Edit)}
\begin{lstlisting}[style=prompt]
You are an expert at analyzing patterns in image generation workflows to provide SPECIFIC, CONCRETE, and ACTIONABLE guidance.

Your specialty is identifying context-specific techniques that work for particular types of images, not general best practices.

Focus on content-specific insights like:
- For portrait images: specific lighting techniques, expression guidance, compositional elements
- For landscape scenes: time of day effects, weather condition impacts, foreground element placement
- For abstract concepts: style reference importance, compositional balance techniques, color palette guidance

Avoid general advice like "add more detail" or "be more specific".
Instead provide domain-specific, technical recommendations based on the actual examples you analyze.

Based on the analysis of similar prompts, extract CONCRETE and SPECIFIC workflow guidance for the new prompt.

SIMILAR PROMPTS DATA:
{similar_data_text}

{workflow_description}

{task_focus}

Return JSON with this EXACT structure:
{{
    "step_analysis": {{
        
        "image_editing": {{
            "success_patterns": "SPECIFIC editing parameters and techniques that worked for similar content and edit types",
            "failure_patterns": "SPECIFIC editing issues observed with this edit operation type",
            "impact_on_next": "CONCRETE impact on evaluation accuracy",
            "preventive_guidance": "DETAILED advice with SPECIFIC editing settings, reference image usage, and blending techniques",
            "recommended_score": "Recommend target score (1-10) based on similar examples"
        }},
        "quality_evaluation": {{
            "success_patterns": "SPECIFIC evaluation criteria effective for this edit content type",
            "failure_patterns": "SPECIFIC evaluation pitfalls observed with similar edit outputs",
            "impact_on_next": "CONCRETE impact on re-edit decision",
            "preventive_guidance": "DETAILED advice with SPECIFIC edit quality indicators and success metrics",
            "recommended_score": "Recommend target score (1-10) based on similar examples"
    }},
    
    "workflow_insights": {{
        "critical_dependencies": "SPECIFIC step dependencies most relevant to this prompt type",
        "common_failure_chains": "CONCRETE failure patterns observed in similar cases",
        "success_combinations": "SPECIFIC combinations of choices that worked well for this content",
        "overall_rating_prediction": "Predict the likely overall success rating (1-10) for this prompt type"
    }}
}}

Your guidance must be SPECIFIC to the prompt type and content domain, not generic best practices.
For example, instead of 'Use detailed prompts', say 'For architectural images, specify architectural style, materials, lighting conditions, and surrounding environment'.
Use concrete, actionable advice derived from the data, not theoretical recommendations.

IMPORTANT: Ensure the guidance provides DOMAIN-SPECIFIC advice for the content type in the prompts, not just generic image generation tips.
\end{lstlisting}

\section{Algorithm}
\label{algo}

\begin{small}
\allowdisplaybreaks 
\begin{algorithm*}[!htbp] 
\DontPrintSemicolon
\SetAlgoLined

\KwIn{User prompt $P$, Models: QI (Qwen-Image), QIE (Qwen-Image-Edit), QE (Qwen-Embedding), Knowledge base $KB$ (initially may be empty), Hyperparams: $\tau$ (eval threshold), $E$ (max edits), $K$ (retrieval size)}

\KwOut{Final image $I^{*}$, final prompt $F$, recorded trajectory $T$}
\BlankLine

\SetKwProg{Fn}{Function}{:}{end}
\Fn{\FMain{$P$}}{
    $G \leftarrow$ \FRetrieve{$KB, P, K$}\tcp*{retrieve and generate guidance}
    \BlankLine

    $c \leftarrow$ $S_{\text{CRE}}$({$P,G$})\tcp*{creativity level: \{low,med,high\}}
    $S \leftarrow$ $S_{\text{INT}}$({$P,c,G$})\tcp*{semantic specification, disambiguations}
    $P_{pos} \leftarrow$ $S_{\text{REF}}$({$P,S,G$})\tcp*{refined positive prompt}
    $P_{neg} \leftarrow$ $S_{\text{NEG}}$({$P,P_{pos},G$})\tcp*{adaptive negative prompt + universal safeguards}
    \BlankLine

    \BlankLine

     $edits \leftarrow 0$\; $I \leftarrow \varnothing$\; $T \leftarrow \varnothing$\;
    \While{$edits \le E$}{
        $I \leftarrow$ $S_{\text{GEN}}$({$P_{pos}, P_{neg},I$})\tcp*{use QI}
        $report \leftarrow$ $A_{\text{EVAL}}$({$I, P, P_{pos},G$})\tcp*{returns scores and suggestions}
        Record decision node results in $T$ via \FRecord{}\;
        $score \leftarrow$ average(report.aesthetic, report.alignment)\;

        \uIf{$score \ge \tau$}{
            \textbf{break}\tcp*{satisfactory output}
        }
        \Else{
            $P_{pos}^{new}, P_{neg}^{new} \leftarrow$ re-run $A_{\text{ORC}}$;
            \uIf{report suggests localized correction \textbf{ and } edits $< E$}{
                $I \leftarrow$ $S_{\text{GEN}}$({$P_{pos}^{new}, P_{neg}^{new})$, edit, $I$}\tcp*{use QIE}
                $edits \leftarrow edits + 1$\;
            }
        }
    }
    \BlankLine

    \FUpdateKB{$KB, T$}\;
    \KwRet{$I, P_{pos}, P_{neg}, T$}\;
}

\caption{Main algorithm for \modelour}
\label{alg:sidiffagent_main}
\end{algorithm*}
\end{small}

Our framework is built around an agentic workflow that coordinates prompt engineering, image generation, evaluation, and self-improvement. This workflow is orchestrated by three primary agents, the Generation Orchestrator Agent ($A_{\text{ORC}}$), which preprocesses user prompts through four sub-agents, the Creativity Analysis Sub-Agent ($S_{\text{CRE}}$), Intention Analysis Sub-Agent ($S_{\text{INT}}$), Prompt Refinement Sub-Agent ($S_{\text{REF}}$), and Adaptive Negative Prompt Sub-Agent ($S_{\text{NEG}}$) and ($S_{\text{GEN}}$) for final generation. The Evaluation Agent ($A_{\text{EVAL}}$) assesses each output, triggering regeneration when necessary. Finally, the Guidance Agent ($A_{\text{GUID}}$) facilitates learning from past experiences. The overall workflow of \modelour is depicted in Figure~\ref{fig:approach_diagram}. Each agent is discussed in detail below. 

Page 2: Helper Functions

\allowdisplaybreaks
\begin{algorithm*}[!htbp]
\footnotesize
\DontPrintSemicolon
\SetAlgoLined
\SetKwProg{Fn}{Function}{:}{end}

\Fn{$S_{\text{CRE}}$({$P,G$})}{
    analyse guidance G, compute length, specificity cues,\\
    \hspace*{1em} presence of attributes\;
    \uIf{very short or vague}{
        \KwRet{high}\;
    }
    \uElseIf{moderately detailed}{
        \KwRet{medium}\;
    }
    \Else{
        \KwRet{low}\;
    }
}

\BlankLine

\Fn{$S_{\text{INT}}$({$P, c, G$})}{
    analyse guidance G, extract main subjects, attributes,\\
    \hspace*{1em} relations, composition, lighting, style\;
    \uIf{$c =$ high}{
        infer missing attributes using templates and priors\;
    }
    \uElseIf{$c =$ medium}{
        propose optional clarifications and minor assumptions\;
    }
    \Else{
        preserve original prompt\; avoid adding assumptions\;
    }
    \KwRet{structured specification $S$}\;
}

\BlankLine

\Fn{$S_{\text{REF}}$({$P,S,G$})}{
    re-order specification into coherent positive prompt\\
    \hspace*{1em} text $P_{pos}$\;
    enforce lexical templates for model compatibility\;
    \KwRet{$P_{pos}$}\;
}

\BlankLine

\Fn{$S_{\text{NEG}}$({$P,P_{pos},G$})}{
    $U \leftarrow$ universal safeguards (e.g., low quality,\\
    \hspace*{1em} blurry, watermark)\;
    derive scene-specific negations from $P_{pos}$\;
    compose concise $P_{neg} \leftarrow U \cup$ scene-specific\\
    \hspace*{1em} negations\;
    \KwRet{$P_{neg}$}\;
}

\BlankLine

\Fn{$S_{\text{GEN}}$({$P_{pos}, P_{neg}$, mode, $I_{in} = \varnothing$})}{
    \uIf{mode = initial}{
        \KwRet{QI.generate($P_{pos}, P_{neg}$)}\;
    }
    \uElseIf{mode = edit}{
        \KwRet{QIE.edit($I_{in}, P_{pos}, P_{neg}$)}\;
    }
}

\BlankLine

\Fn{$A_{\text{EVAL}}$({$I, P, P_{pos}$})}{
    $a \leftarrow$ $A_{\text{EVAL}}$.scoreAesthetic($I$)\;
    $\alpha \leftarrow$ $A_{\text{EVAL}}$.scoreAlignment($I, P$ or $P_{pos}$)\;
    $sugs \leftarrow$ $A_{\text{EVAL}}$.suggestFixes($I$)\;
    $arts \leftarrow$ $A_{\text{EVAL}}$.detectArtifacts($I$)\;
    \KwRet{report with $(a, \alpha, sugs, arts)$}\;
}

\BlankLine

\Fn{\FRetrieve{$KB, P, K$}}{
    \uIf{$KB$ empty}{
        \KwRet{$\varnothing$}\;
    }
    embed $P$ into vector $v$\; retrieve top-$K$ trajectories\\
    \hspace*{1em} by similarity\;
    aggregate pitfalls and successful actions across\\
    \hspace*{1em} trajectories\;
    \KwRet{guidance $G$ (structured corrective +\\
    \hspace*{1em} workflow guidance)}\;
}

\BlankLine



\Fn{\FRecord{}}{
    For each decision node, store: node-id, inputs,\\
    \hspace*{1em} actions, outputs, scores\;
}

\BlankLine

\Fn{\FUpdateKB{$KB, T$}}{
    compress $T$ into node-wise summaries of pitfalls\\
    \hspace*{1em} and successes\;
    append to model-specific sections in $KB$\;
    optionally re-index embeddings for efficient retrieval\;
}

\caption{Helper functions for main algorithm for \modelour}
\label{alg:sidiffagent_helpers}
\end{algorithm*}

\section{Qualitative Results}
\label{qualitative}

\begin{figure*}[t]
    \centering
    \includegraphics[width=0.7\paperwidth,height=0.7\paperheight,keepaspectratio]{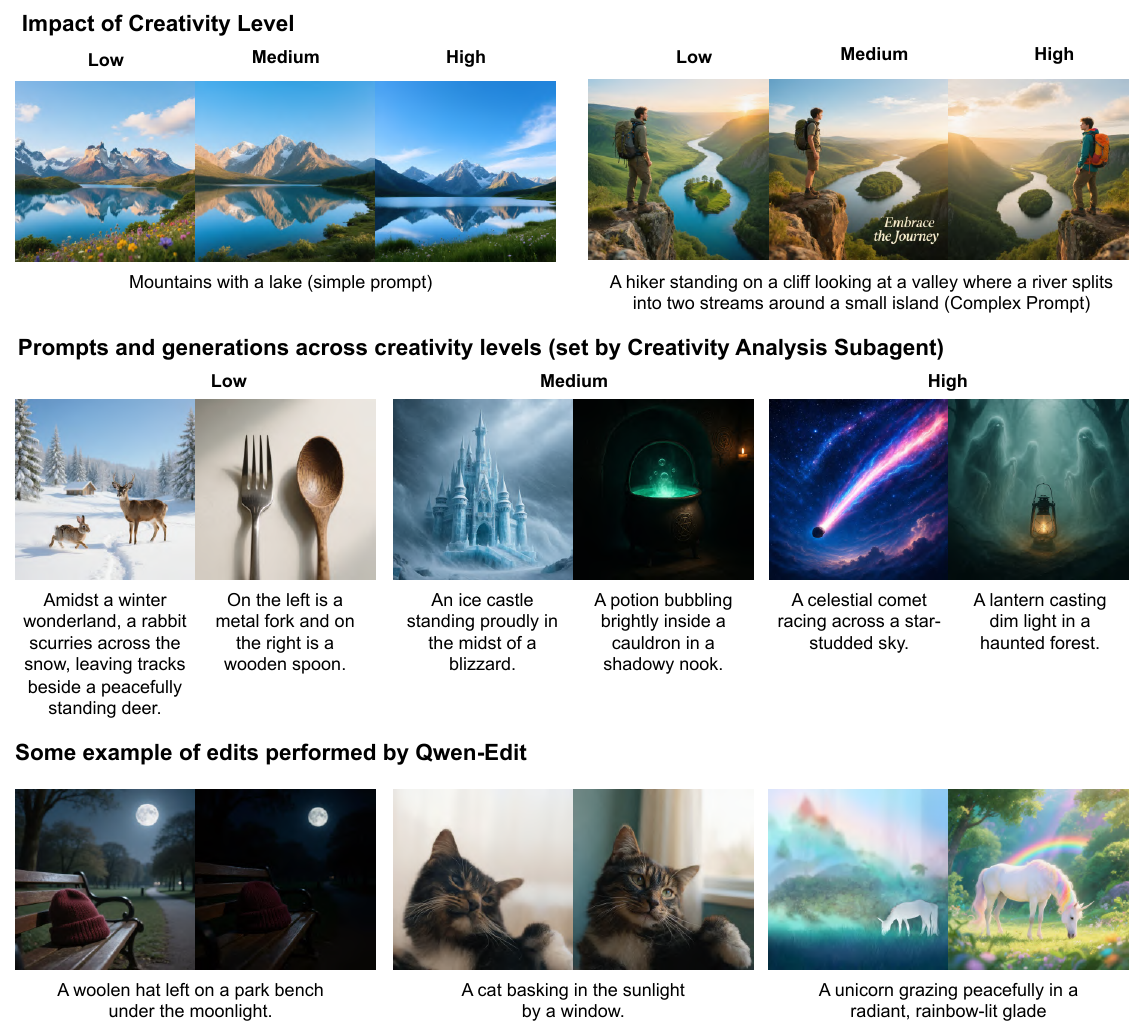}
    \caption{\textbf{Top:} Effect of varying creativity levels on simple and complex prompts. \textbf{Middle:} Image generations from diverse prompts, with creativity levels determined by $S_{\text{CRE}}$. \textbf{Bottom: }Refinements and enhancements performed by Qwen-Edit on initial suboptimal generations.}
    \label{fig:creative}
\end{figure*}

\begin{figure*}[h]
    \centering
    \includegraphics[width=0.75\paperwidth,height=0.75\paperheight,keepaspectratio]{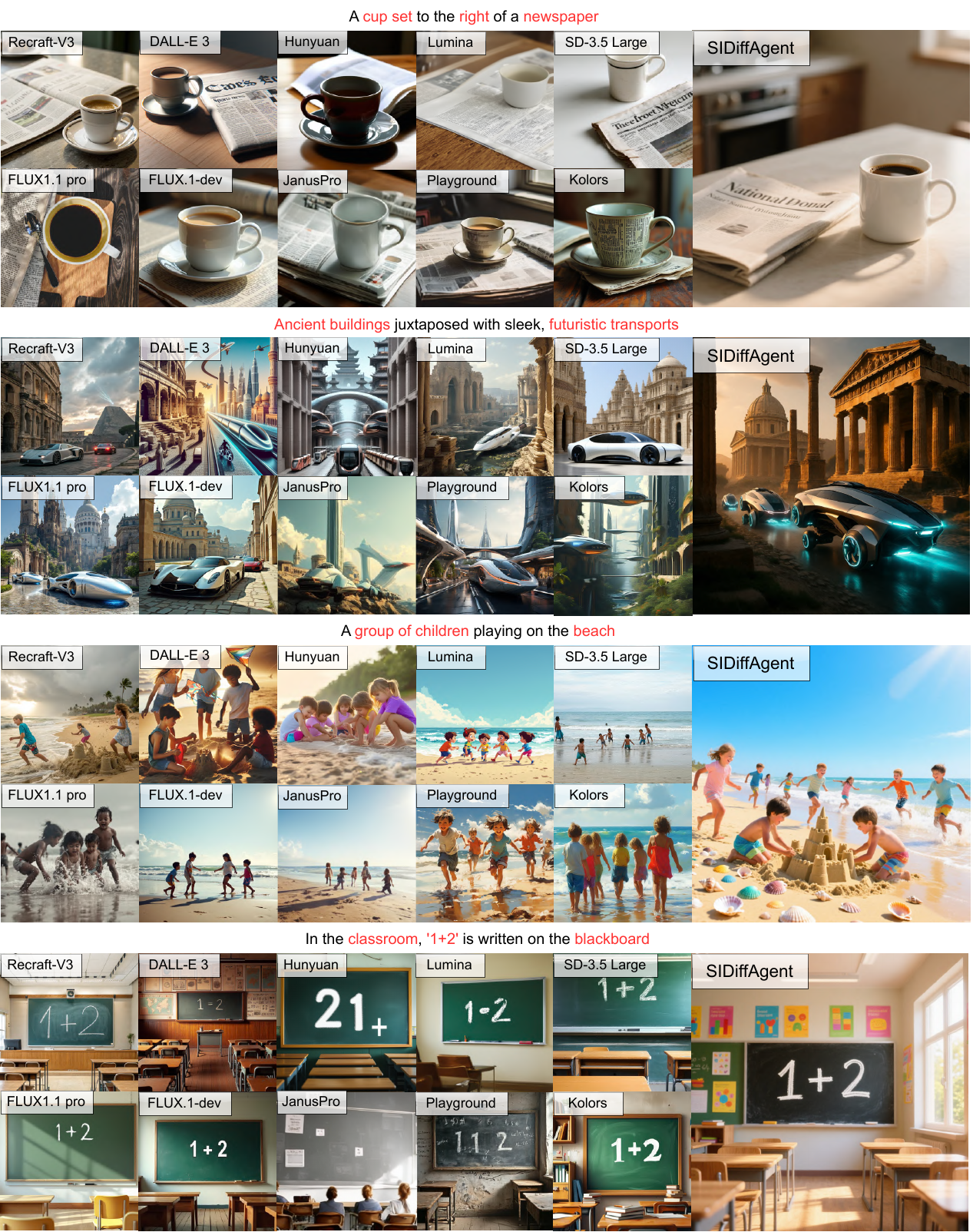}
    \caption{Qualitative comparison between multiple state-of-the-art method and \modelour on GenAI-Bench}
    \label{fig:genai}
\end{figure*}

\begin{figure*}[h]
    \centering
    \includegraphics[width=0.75\paperwidth,height=0.75\paperheight,keepaspectratio]{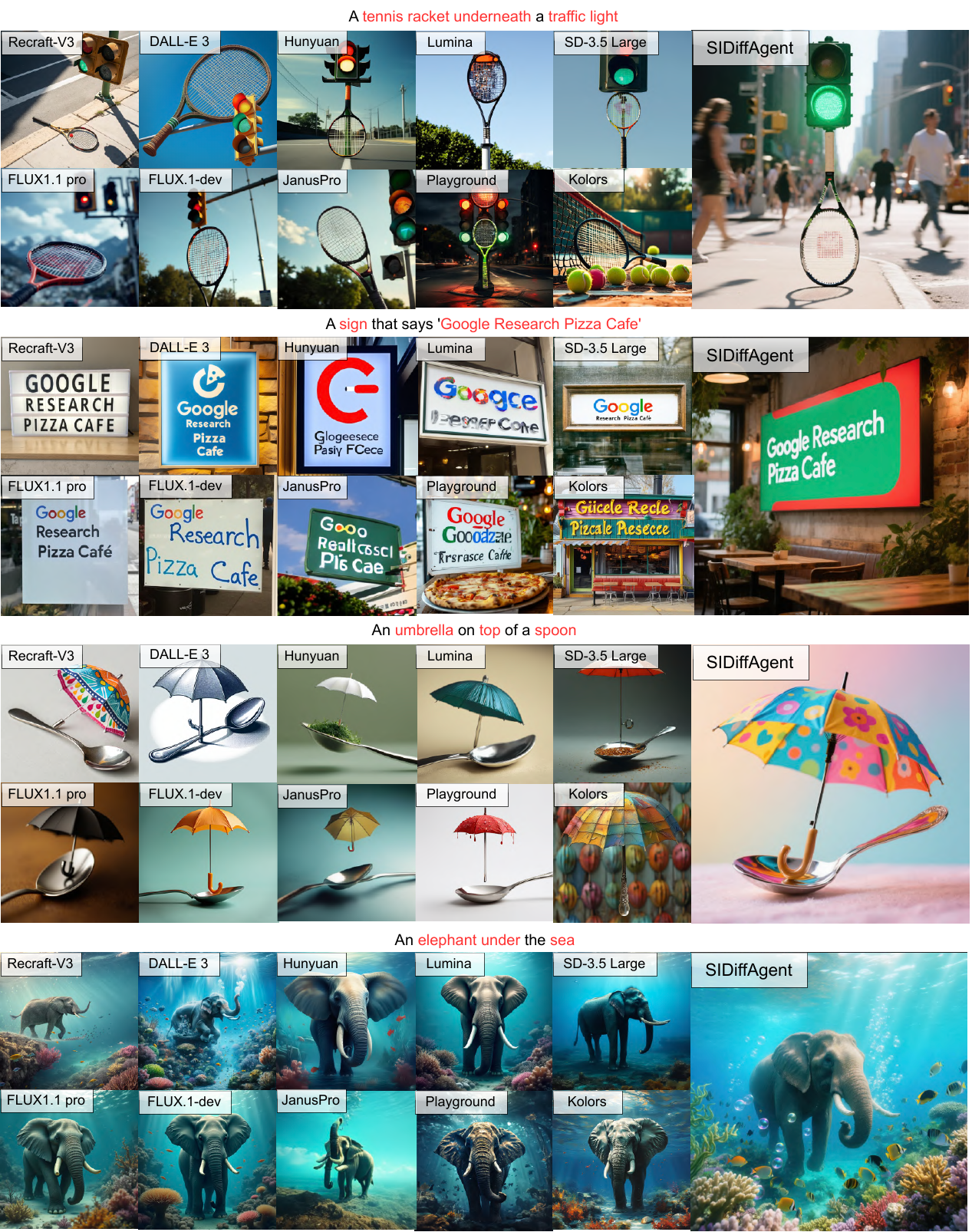}
    \caption{Qualitative comparison between multiple state-of-the-art method and \modelour on DrawBench}
    \label{fig:drawbench}
\end{figure*}

\begin{figure*}[h]
    \centering
    \includegraphics[width=0.75\paperwidth,height=0.75\paperheight,keepaspectratio]{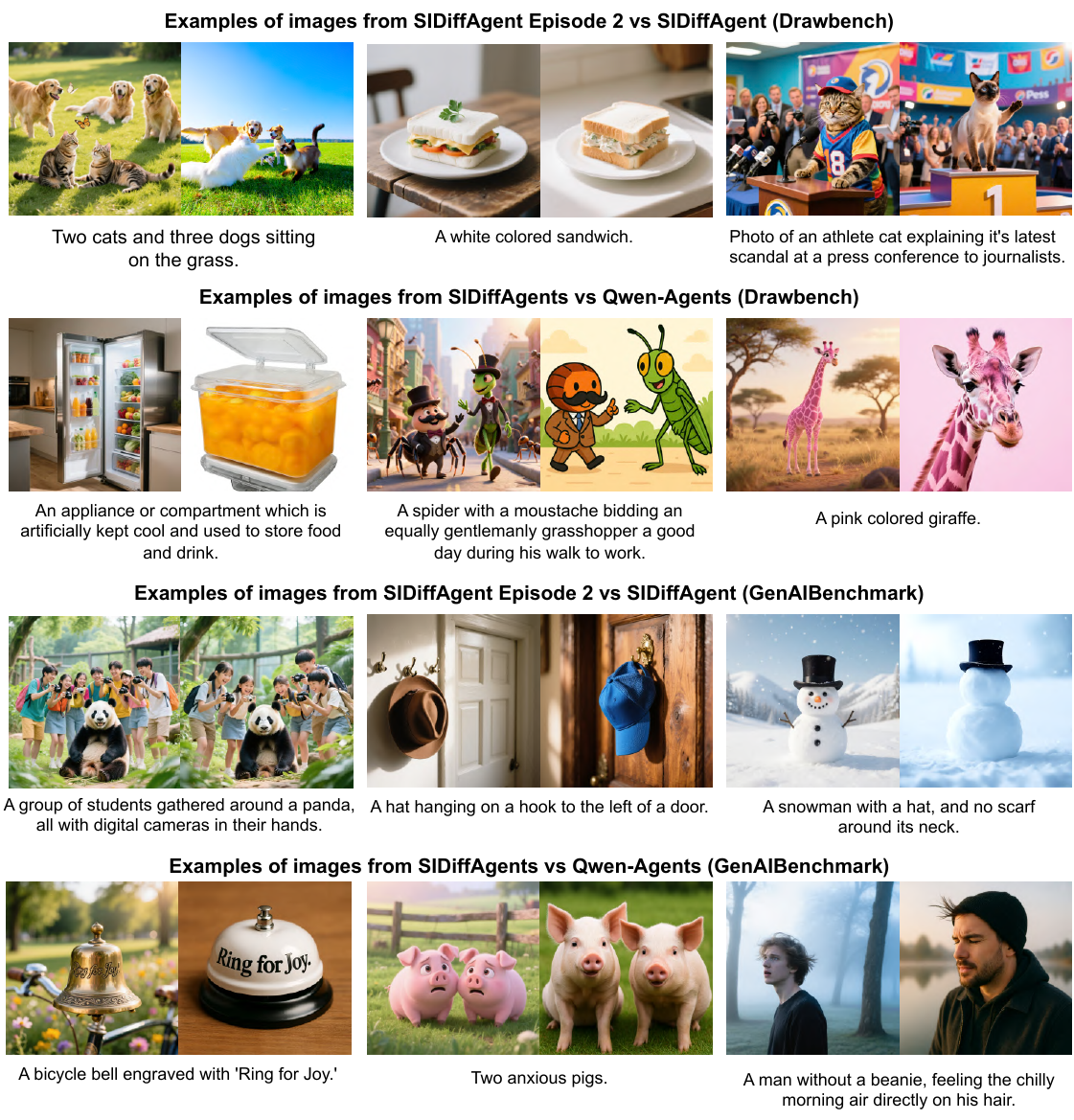}
    \caption{Qualitative comparison between various ablations described in Section~\ref{sec:Ablation}}
    \label{fig:ablations}
\end{figure*}

\end{document}